\documentclass{article}
\usepackage[utf8]{inputenc}

\usepackage[square,sort&compress,comma,numbers]{natbib}
\usepackage{hyperref}
\usepackage{xcolor}
\usepackage[letterpaper,top=2cm,bottom=2cm,left=2cm,right=2cm,marginparwidth=1.75cm]{geometry}

\usepackage{palatino,graphicx,wrapfig}
\usepackage[small,it]{caption}
\usepackage{amsmath}
\usepackage{amssymb,bm}
\usepackage{parskip}
\usepackage{authblk}

\title{Multiscale Graph Neural Network Autoencoders for Interpretable Scientific Machine Learning}
\author[1]{Shivam Barwey}
\author[2]{Varun Shankar}
\author[2]{Venkatasubramanian Viswanathan}
\author[3]{Romit Maulik}

\affil[1]{Argonne Leadership Computing Facility, Argonne National Laboratory, USA}
\affil[2]{Department of Mechanical Engineering, Carnegie Mellon University, USA}
\affil[3]{Mathematics \& Computer Science Division, Argonne National Laboratory, USA}

\begin{document}
\date{}
\maketitle

\begin{abstract}
The goal of this work is to address two limitations in autoencoder-based models: latent space interpretability and compatibility with unstructured meshes. This is accomplished here with the development of a novel graph neural network (GNN) autoencoding architecture with demonstrations on complex fluid flow applications. To address the first goal of interpretability, the GNN autoencoder achieves reduction in the number nodes in the encoding stage through an adaptive graph reduction procedure. This reduction procedure essentially amounts to flowfield-conditioned node sampling and sensor identification, and produces interpretable latent graph representations tailored to the flowfield reconstruction task in the form of so-called masked fields. These masked fields allow the user to (a) visualize where in physical space a given latent graph is active, and (b) interpret the time-evolution of the latent graph connectivity in accordance with the time-evolution of unsteady flow features (e.g. recirculation zones, shear layers) in the domain. To address the goal of unstructured mesh compatibility, the autoencoding architecture utilizes a series of multi-scale message passing (MMP) layers, each of which models information exchange among node neighborhoods at various lengthscales. The MMP layer, which augments standard single-scale message passing with learnable coarsening operations, allows the decoder to more efficiently reconstruct the flowfield from the identified regions in the masked fields. Analysis of latent graphs produced by the autoencoder for various model settings are conducted using unstructured snapshot data sourced from large-eddy simulations in a backward-facing step (BFS) flow configuration with an \verb|OpenFOAM|-based flow solver at high Reynolds numbers. 
\end{abstract}
\tableofcontents

\section{Introduction}
\label{sec:intro}
Computational fluid dynamics (CFD) plays a key role in guiding both the design and physical understanding of a variety of applications -- these include, among others, the exploration of novel fuel-efficient propulsion concepts \cite{raman2019emerging}, the prediction of pollutant dispersion in urban environments \cite{blocken2012ten}, modeling the evolution of wildfire spread \cite{bakhshaii2019review}, and accelerating the design of wind turbine farms \cite{sanderse2011review}. The flowfields that govern many of these applications can only be described accurately using partial differential equations (PDEs) that model multi-physics interactions driven by the coupling between large scale and small scale transient phenomena. Ultimately, as these applications become more complex and the available computing power continues to increase at unprecedented levels, the amount of available data in the form of high-quality flowfields will inevitably increase as well \cite{exascale_apps}. 

The impact in the past decade in the fluid dynamics community has been the development and maturation of a wide variety of techniques that necessarily scale with the increase in available data (i.e. modeling frameworks that thrive in data-rich environments) \cite{kochkov2021machine,karthik_arfm,raman2019emerging}. These data-based modeling approaches often apply order reduction strategies to (a) accelerate conventional CFD-based simulations using projection-based reduced-order modeling (ROM) as alternatives to statistical approaches (e.g. large-eddy simulations), (b) replace conventional CFD by discovering predictive computational models from experimental or real-world data directly, and (c) extract salient features from complex flowfields in both numerical and experimental settings to assist in expert-guided physical understanding for downstream model development. 

Token examples of data-based modeling strategies come from the class of modal decomposition methods \cite{taira_review_aiaa}, such as proper orthogonal decomposition (POD) \cite{berkooz_pod}, dynamic mode decomposition (DMD) \cite{schmid_2010}, and data clustering approaches \cite{kaiser_jfm}. Here, order reduction is achieved through a factorization of a snapshot representation of the fluid flow evolution into a small set of spatial modes. The modes are then used to develop a ROM via linear projection onto the governing PDEs. State-of-the-art implementations generalize these modal decomposition approaches by utilizing the expressive power of neural networks to develop methods for field transformations (i.e. recovering velocity fields fields from scalar fields \cite{dahm_1996,shivam_cst}), super-resolution \cite{hassanaly2022adversarial,fukami2019super}, and autoencoder-based compression \cite{romit_pof_2021}. In the context of model development, the autoencoding goal is to replace linear projection used in methods like POD with non-linear projection that improves the predictive accuracy of unsteady processes. Autoencoding architectures for ROM development consist of an encoder-integrator-decoder procedure, where the encoder and decoder are cast as neural networks \cite{romit_pof_2021,karthik_cnn_ae,carlberg_nonlinear}. In a broad sense, the encoder is a type of projection operator that transforms a high-dimensional instantaneous flowfield -- the initial condition supplied to the CFD solver, for example -- into a latent space that compresses this flowfield into a low-dimensional representation. Then, the integrator (the prognostic model) solves a modified set of ODEs in the latent space to effectively advance the low-dimensional representation forward in time, leading to significant computational savings (the integrator can be physics-based, or can leverage other data-based strategies such as neural ordinary differential equations \cite{romit_pod_2020,chen2018neural}, Markovian transition matrices \cite{kaiser_jfm,shivam_ctm}, or recurrent neural networks \cite{romit_spod_2022,reddy2019reduced}). The decoder is then used to recover the time-advanced high-dimensional flowfield after the integrator stage, such that the end-result can be interpreted and visualized by field experts for physical analysis or for feeding into flowfield-based actuation/mitigation strategies. The crux of the approach lies in the manner in which the latent space is created -- the interpretability, physical significance, and generalizability of the latent variables depend on the type of autoencoding strategy used. For example, popular approaches include multi-layer perceptrons (MLPs), orthogonal basis projections via POD \cite{romit_pod_2020}, or convolutional neural networks \cite{murata2020nonlinear}. 

Although the overall success of the above data-based modeling strategies is inarguable when it comes to capturing complex physics, some key limitations prohibit their extensions into practical flow configurations. Many realistic fluid flow applications are described by complex geometries (propulsion devices with various injection schemes, airfoil geometries, wind turbines, etc.) -- key requirements for data-based ROM development in such scenarios are (a) ensuring inherent compatibility with flowfields stored on unstructured grids compatible with these complex geometries, and (b) accounting for the fact that variations in geometric configurations should not require additional highly expensive training stages. When considering both of these requirements, many of the aforementioned data-based ROM approaches break down. For example, POD-based models recovered from a single geometric configuration cannot be reliably extended to other configurations. Autoencoders based on convolutional neural networks (CNNs) are by design restricted to structured representations of the flowfield data. Similarly, ROM strategies reliant on MLP-based autoencoders are also often restricted to single geometric configurations, and the neural network parameters are obtained in such a way that does not allow for inference on unexplored or unseen meshes. Another limitation is model interpretability -- in many studies, data-based ROM tools are essentially black boxes, rendering latent spaces uninterpretable. Ensuring interpretability in modeling frameworks often leads to (a) simplification and reductions in network complexity or expressive power (e.g. using a linear projection basis instead of a nonlinear one, for example \cite{romit_pod_2020,shivam_proci}), and/or (b) the development of other optimization strategies geared towards correlating the discovered latent dynamics with physical processes (i.e. disentanglement of latent variables \cite{chen2018isolating,karthik_disentanglement_2022}), which adds additional computational expense and training constraints. Ultimately, the need for interpretation of the internal structures found in neural networks -- particularly for applications like fluid flow prediction -- has led to high modern demands for explainable artificial intelligence \cite{xai}. 

The goal of this work is to address both of these limitations -- namely, interpretability and compatibility with unstructured meshes -- by means of leveraging graph neural network (GNN) architectures for data-based model development. Within this scope, the main objective of this work is to show how the development of a novel graph autoencoder leads to highly interpretable latent representations. Before proceeding, some additional background and review on graph neural networks in the context of fluid dynamics modeling is provided below, as is a more concrete description of the individual contributions of this work. 

Introduced in early works in Ref.~\cite{sperduti_1997,gori_2005}, graph neural networks have gained popularity in recent years due to their ability to combine the scalability of backpropagation-based optimization with flexible representations of data \cite{kipf_gcn,xu2018powerful}. The end-result is a framework for executing generalized classification and/or regression tasks on any class of problems that can be cast as a complex network or graph. The generalizability and ambiguity in graph representations of data lends to their overall strength and increasing usage in most modeling tasks: so long as a problem or dataset can be described as a set of nodes, with connections (edges) among nodes encoding some notion of relational distance or feature similarity, a GNN-based approach can be utilized  \cite{battaglia_2018}. To this end, modeling strategies based on GNNs leveraging message passing \cite{gilmer_2017} and graph Laplacian \cite{defferrard_2016} operations have produced state-of-the-art results on image classification, object detection, and node classification tasks. Additionally, due to their versatility, GNNs see significant success in multi-disciplinary applications including modeling protein folding \cite{graph_protein}, predicting the emergent properties of social networks \cite{graph_social}, and modeling differential equations \cite{grand,graph_node}. 

Much of the success in these pioneering GNN applications has translated into new data-based model development strategies for fluid dynamics applications, with the primary goal of providing a mesh-agnostic way to accelerate conventional CFD flow solvers. Because CFD simulations are stored on mesh arrangements from the get-go with pre-defined stencils (node connectivities), a data-based flowfield regression framework fits naturally for these problems within the GNN paradigm. The key idea is that standard message passing strategies used within GNN regression frameworks are dependent on not only node features in isolation (node features can be pressure, velocity, or any other flow quantity of interest), but also the properties of the neighborhood of the nodes in question. Due to this property, if the input CFD mesh is considered as a set of points from which a graph is derived, several applications have shown how a GNN trained on one mesh can be readily extended to other meshes (or other geometries) without the need for re-training. 

With this capability in mind, GNNs for fluid flow modeling have been deployed in a variety of contexts for both steady-state \cite{gcn_cfd_differentiable,peng2022grid,yang2022amgnet} and unsteady problems \cite{deepmind_2020,lino2022multi}. In steady-state applications, GNNs are typically used to reconstruct flowfields from geometry-dependent fields such as signed distance functions. For example, Ref.~\cite{gcn_cfd_differentiable} creates a surrogate solver for steady-state flows using a super-resolution strategy based on the graph convolution network \cite{kipf_gcn}. Additional analysis into the role of graph pooling operations in the reconstruction procedure for steady-state flowfields was performed in Ref.~\cite{graph_cfd_benchmark}. From the perspective of unsteady fluid flow prediction, a key example of GNN-based modeling is the MeshGraphNet architecture introduced in Ref.~\cite{deepmind_2020}, in which an encode-process-decode strategy is used to learn the source terms as spatial fields on unstructured grids for unsteady incompressible flow over cylinders and airfoils, among other applications. Inspired by methods used in multigrid methods \cite{mavriplis1990multigrid}, multi-scale GNN archtitectures (a variant of which is leveraged in this work) have recently emerged to address limitations in single-scale counterparts by levaraging graph coarsening operations \cite{multiscale_meshgraphnet,lino2022multi,yang2022amgnet,cao2022bistride}.

Within the context of fluid flow prediction, although the above GNN-based predictive models succeed in extending conventional neural networks to unstructured grids and variable geometries, there are some lingering limitations. For example, most existing approaches leverage graph connectivities that are fixed in time. There has been very little emphasis on utilizing learnable, adaptive graph pooling strategies within the GNN architecture -- these adaptive pooling layers are essentially operations which condition the generation of coarse graph connectivities (adjacency matrices) on some input field stored on the nodes of a fine, or baseline, graph. Although adaptive pooling strategies have been recently explored \cite{yang2022amgnet}, these applications are typically restricted to steady-state flows and utilize dense interpolation operators that cannot scale to large graph sizes. Further, much of the recent focus in GNNs has been geared towards purely forecasting operations -- in other words, given a flowfield at some time $t$, the objective is to learn the evolution at some future time $t + \Delta t$. Although the direct forecasting task is indeed relevant, efforts into purely graph-based autoencoding (i.e. using a GNN to recover an identity map) have been sparsely explored. It is noted that recent work in Ref.~\cite{gunzburger_2022} used graph convolution operations in an autoencoding context to good effect; however, the methods presented in this work are distinct in that the architecture is completely based on graphs, incorporating graph pooling and unpooling layers without any flattening operations.

The goal of this work is to address the above described limitations and gaps in the existing GNN literature for unsteady fluid flow modeling. The specific contributions are as follows: 
\begin{itemize}
    \item A GNN-based autoencoding architecture is developed with the purpose of ensuring latent space interpretability. This is accomplished by employing an adaptive pooling strategy known as Top-K pooling \cite{gao2019graph}. The Top-K pooling mechanism is considered as a type of a node sampling procedure, or adaptive sensing strategy, used to reduce the total number of nodes needed to minimize the target objective. The end-result is a latent space, or bottleneck layer, that can can be visualized in physical space directly, identifies coherent structures in the domain, and is described by an adjacency matrix that adapts in time with the evolution of the flow. 
    \item A series of multi-scale message passing (MMP) layers are employed between pooling operations, each of which models information exchange among node neighborhoods at various lengthscales. The MMP layer, which augments standard single-scale message passing with learnable coarsening operations, allows the decoder to more efficiently reconstruct the flowfield from the identified regions in the latent graphs. 
    \item The GNN autoencoder is demonstrated on CFD snapshot data sourced from unsteady flow over a backward-facing step at Reynolds numbers in the range of roughly $20000$ to $50000$. Analysis of latent spaces are conducted on these snapshots to demonstrate (a) interpretability and physical significance of latent spaces via masked field visualization, and (b) trade-offs between reconstruction accuracy, interpretability, and achievable compression produced by the latent graphs. 
\end{itemize}
The remainder of the text proceeds as follows. The dataset generation procedure, governing equations, and domain geometry are described in Sec.~\ref{sec:config_data}. An overview of the methodology is then provided in Sec.~\ref{sec:methodology}, which describes the graph generation procedure and architecture description. Autoencoding results, with emphasis on analysis of the interpretable Top-K pooling output, is then provided in Sec.~\ref{sec:results}. Concluding remarks follow in Sec.~\ref{sec:conclusion}.  

\section{Configuration and Dataset}
\label{sec:config_data}
\subsection{Governing equations and flow solver}
\label{sec:gov_eq}
The data used in this study comes from unsteady, 2-dimensional large-eddy simulations (LES) of the incompressible Navier-Stokes equations of flow over a backward-facing step (BFS). Before proceeding, it should be noted that the application of LES in this scenario as a modeling approach is not entirely physical due to the usage of a 2-dimensional configuration; rather, turbulence modeling was applied here as a mechanism to produce coherent structures in flowfields that are reasonably complex and unsteady in nature for GNN demonstration purposes. As such, efforts to tune the LES model for physical consistency (i.e. consideration of near-wall regions) were not carried out.

More specifically, the governing equations in the LES formulation are, from the conservation of mass and momentum in incompressible flow \cite{popebook},
\begin{equation}
    \label{eq:gov_eq}
    \begin{split}
    \frac{\partial \widetilde{u}_j}{\partial t} + \frac{\partial \widetilde{u}_i \widetilde{u}_j}{\partial x_i} &= \nu \frac{\partial^2 \widetilde{u}_j}{\partial x_i \partial x_i} - \frac{\partial \tau_{ij}^r}{\partial x_i} - \frac{1}{\rho}\frac{\partial \widetilde{p}}{\partial x_j}, \\
    \frac{\partial \widetilde{u}_k}{\partial x_k} &= 0.
    \end{split}
\end{equation}
For the general time-evolving field $f(x,t)$, the spatial filtering operation, denoted $\widetilde{.}$, is defined as the convolution
\begin{equation}
    \widetilde{f}(x) = \int f(x',t) G(x, x') dx', 
\end{equation}
where the kernel $G$ is called the filter. In this work, the filtering is performed implicitly -- the flow variables stored on the computational grid are interpreted as the output of a box-filtering operation of filter size specified by the grid resolution \cite{popebook}. In Eq.~\ref{eq:gov_eq}, the quantity $\widetilde{u}_j = \widetilde{u}_j(x,t)$ denotes the time-evolving $j$-th component of the filtered velocity, $\widetilde{p} = \widetilde{p}(x,t)$ is the time-evolving filtered pressure modified to include the isotropic component of the residual stress tensor, $\nu$ is a constant kinematic viscosity, $\rho$ is a constant density, and $\tau^r = \tau^r(x,t)$ is the deviatoric component of the residual, or sub-grid scale (SGS), stress tensor, defined as 
\begin{equation}
    \label{eq:stress_tensor} 
    \tau_{ij}^r = \tau_{ij}^R - \frac{1}{3} \tau_{ii}^R \delta_{ij}, 
\end{equation}
where $\tau_{ij}^R = \widetilde{u_i u_j} - \widetilde{u_i}\widetilde{u_j}$. Because the residual stress tensor in Eq.~\ref{eq:stress_tensor} cannot be determined due to the presence of the filtered nonlinear term $\widetilde{u_i u_j}$, a model is needed close the partial differential equations in Eq.~\ref{eq:gov_eq}. There are many pathways available to this end -- in this work, the standard Smagorinsky model is used \cite{smagorinsky}, which casts the residual stress $\tau_{ij}^r$ as a quantity proportional to the filtered rate-of-strain as
\begin{equation}
    \label{eq:smagorinsky}
    \begin{split}
    \tau_{ij}^r &= -2 \nu_r \widetilde{S}_{ij},\\
    \nu_r &= \left(C_S \Delta \right)^2 |S|.  
    \end{split}
\end{equation}
In Eq.~\ref{eq:smagorinsky}, $S_{ij}$ is the filtered rate-of-strain, $|S| = \sqrt{2S_{ij}S_{ij}}$ is its magnitude, $\nu_r$ is the turbulent eddy-viscosity, $C_S$ is the Smagorinsky constant ($C_S=0.168$ here), and $\Delta$ is the box-filter width.  

The open-source library \verb|OpenFOAM| is used to implement the numerical schemes needed to solve Eq.~\ref{eq:gov_eq} within the finite-volume framework. \verb|OpenFOAM| is also used here to produce the underlying computational mesh of the flow configuration (the mesh and configuration geometry is described in Sec.~\ref{sec:config}). The distribution of the \verb|OpenFOAM| library contains a number of pre-packaged flow solvers used for the purposes of treating unsteady incompressible flows over complex geometries that are also compatible with a set of LES models, one of which is the Smagorinsky model described above. Here, the baseline \verb|pimpleFOAM| solver is used with globally second-order numerics and one outer iteration. The central theme in the \verb|pimpleFOAM| solver is to treat the effects of the pressure gradient term during a single time step with an operator splitting strategy resembling a predictor-corrector approach \cite{piso_1986,simple}. Full details on the numerical schemes are out-of-scope here; the authors instead point the reader to Ref.~\cite{openfoam_fv} for more information on the specifics of \verb|OpenFOAM| numerics. In the end, the user must describe the spatial numerical schemes, temporal schemes, and LES model parameters in a set of input files. These files -- \verb|fvSchemes|, \verb|fvSolution|, and \verb|momentumTransport| respectively -- are provided in the supplementary material. In the remainder of the paper, for notational convenience, the filter notation $\widetilde{.}$ used to describe velocity fields is dropped without loss of generality. 
 
\subsection{Flow configuration and mesh}
\label{sec:config}
\begin{figure}
    \centering
    \includegraphics[width=\columnwidth]{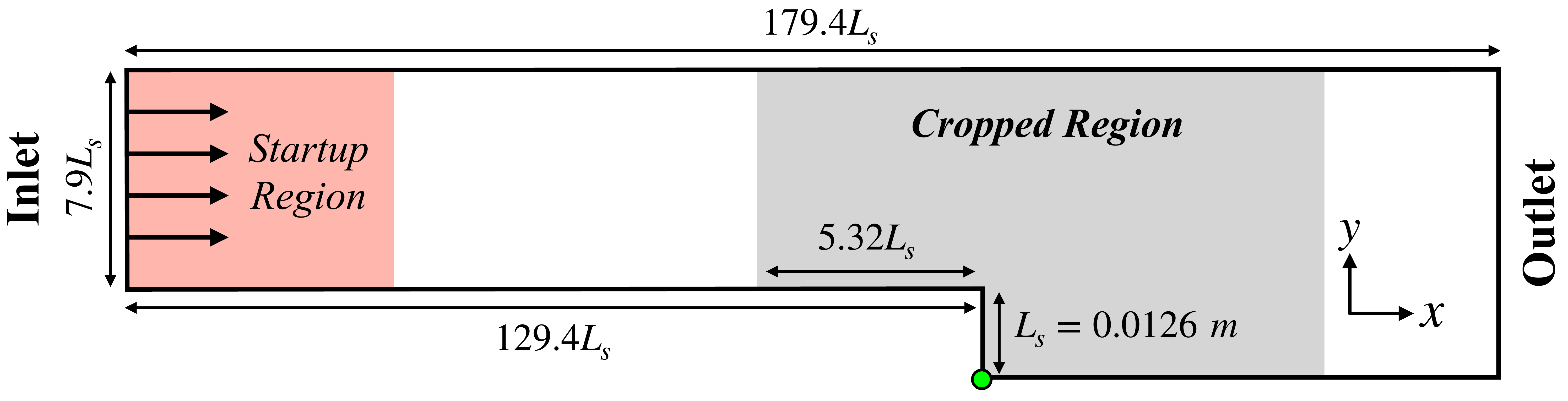}
    \caption{Schematic of the backward-facing step configuration (not shown to scale) using an inlet-to-step length ratio of $7.9$. Lengths are normalized by the step height $L_{s}$, and green circle marks the origin. The mesh for the cropped region indicated in this figure is shown in Fig.~\ref{fig:mesh}. }
    \label{fig:bfs_schematic}
\end{figure}

With the governing equations and flow solver described above, a general schematic of the BFS configuration is shown in Fig.~\ref{fig:bfs_schematic}. This configuration, along with the similar double-BFS (or cavity) configuration, has been used in a number of experimental \cite{armaly1983experimental,kostas2002particle} and numerical \cite{lee1998experimental,wee2004self} studies as a benchmark model problem for flows containing separation and re-attachment phenomena dominated by turbulent mixing, recirculation zones (as well as transition to turbulence), and vortex shedding. Flow enters from an inlet on the left and propagates through an initial channel of fixed width upon encountering a step anchor, triggering flow separation upon entering a second channel of larger width. The step size, $L_s$, is a key parameter in this configuration as it controls primary unsteady flow quantities of interest such as recirculation zone sizes observed in the cavity formed behind the step, vortex shedding frequencies downstream, and shear layer dynamics separating the cavity and freestream region. Values for the geometric quantities used in the simulation procedure are also provided in Fig.~\ref{fig:bfs_schematic}.  
During the simulation procedure, velocity at the inlet is prescribed in the x-direction to satisfy a Reynolds number within the range $[26214, 45589]$, with characteristic length-scale based on the step height $L_{s}$. Additionally, for stability purposes, an initial startup region is used (indicated in red in Fig.~\ref{fig:bfs_schematic}) to provide a uniform velocity profile to the channel. In this startup region, slip boundary conditions are used at the walls; no-slip conditions are used at all other boundaries except for the inlet and outlet.

\begin{figure}
    \centering
    \includegraphics[width=\columnwidth]{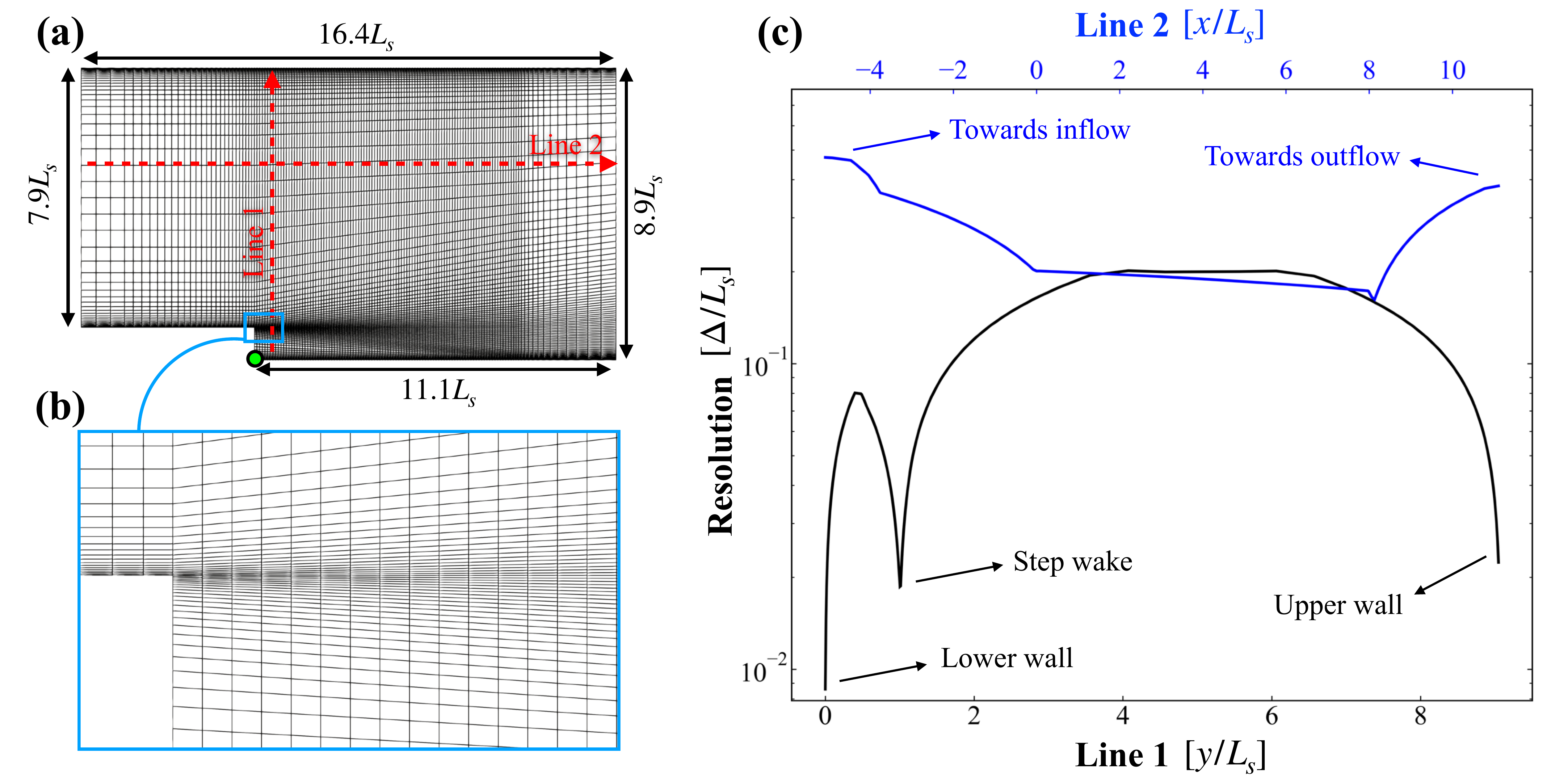}
    \caption{\textbf{(a)} Cropped mesh of the BFS configuration shown to scale. Red lines indicate lines along which resolutions are computed in (c). Green circle denotes origin. \textbf{(b)} Zoom-in of the mesh highlighting refinement near the step. \textbf{(c)} Mesh resolution $\Delta$ normalized by the step height $L_{s}$ along lines shown in (a).}
    \label{fig:mesh}
\end{figure}

Although the flowfield data were obtained from runs in the full BFS geometry, for the purpose of dropping training times, data used to train the GNNs come from a subdomain (the gray cropped region in Fig.~\ref{fig:bfs_schematic}) containing all flow dynamics of interest. Note that no interpolation procedures or mesh changes were utilized when extracting this cropped region -- rather, a subset of cells were selected using the \verb|cellSet| functionality in \verb|OpenFOAM| such that only regions near the step were captured. A visualization of the cropped mesh used in this study is shown in Fig.~\ref{fig:mesh}(a), with a zoom-in on the near-step region in Fig.~\ref{fig:mesh}(b). The mesh, generated using the \verb|blockMesh| utility in \verb|OpenFOAM|, contains 14476 control volumes (cells) and is characterized by highly refined regions near walls and in a region behind the step such that (a) boundary layers are reasonably resolved, and (b) unsteady dynamics in the near-step fan, such as flow separation and emergence of recirculation zones, are properly captured. Mesh resolution increases as the domain progresses to the outlet and exits the near-step region -- this coarsening effects continues outside of the cropped region in the x-direction. This variation in cell resolution $\Delta$ is shown in Fig.~\ref{fig:mesh}(c) across the two perpendicular lines indicated in Fig.~\ref{fig:mesh}(a). The mesh in Fig.~\ref{fig:mesh} contains noticeable variations in cell sizes and skewness, and is therefore a sufficiently practical benchmark to assess GNN capabilities in complex geometric configurations. 

\subsection{Description of dataset}
\label{sec:dataset}

\begin{figure}
    \centering
    \includegraphics[width=\columnwidth]{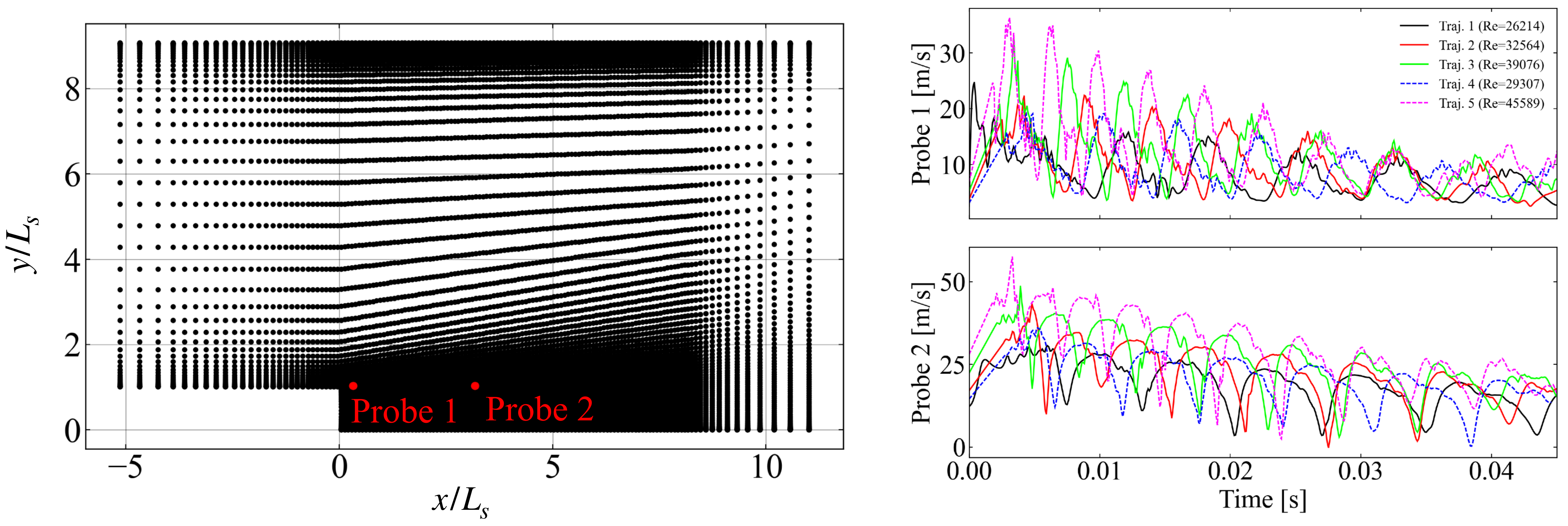}
    \caption{\textbf{(a)} Red markers indicate probe locations in the BFS configuration used to plot trajectories in (b) and (c). Black points mark control volume centroids. \textbf{(b)} Evolution of pressure for probe locations depicted in (a). Data collection phase occurs after roughly $t=0.14$~s. Note that pressure here can be negative because the quantity is shifted by an arbitrary reference. Pressure has also been normalized by the constant density $\rho$. \textbf{(c)} Visualization of training, validation, and testing datasets using pressure evolution at the probe locations in (a). Bold lines isolate single trajectories for clarity.}
    \label{fig:data_collection}
\end{figure}

Using the BFS simulation configuration, datasets used for training and testing the GNN models described in Sec.~\ref{sec:methodology} consist of a collection snapshots describing the evolution two-component velocity fields in the cropped mesh shown in Fig.~\ref{fig:mesh} at various Reynolds numbers. An instantaneous snapshot in this context is given by the matrix ${\bf X}(t) \in \mathbb{R}^{N_C \times N_F}$, where $t$ denotes the time, $N_C$ the number of cells or control volumes ($N_C=14667$), and $N_F$ the number of features ($N_F=2$ here for streamwise and vertical velocity components). The rows of each snapshot ${\bf X}(t)$ describe individual cell centroid locations that are fixed in physical space as provided by the mesh configuration. The columns indicate the flowfield component, or feature, stored at the corresponding cell index in accordance with the finite volume formulation (Fig.~\ref{fig:data_collection} shows a visualization of the cell centers). 

The complete dataset consists of 5 simulation trajectories, each of which comes from sampling flowfields at a fixed time interval of $\Delta t= 10^{-4}$~seconds at the various Reynolds numbers described in Table~\ref{tab:dataset}. Note that although the sampling interval is fixed, the simulations themselves were run in a variable timestep setting based on a maximum Courant-Friedrichs-Lewy (CFL) number of $0.5$. As shown in Table~\ref{tab:dataset}, the training data is comprised 3 out of the 5 total trajectories corresponding to Reynolds numbers of 26214, 32564, and 39076 respectively, producing a total number of $1216$ training snapshots. The testing data consists of the remaining 2 trajectories: one at an interpolated Reynolds number with respect to the training data range (Re=29307), and the other at a larger, extrapolated Reynolds number (Re=45589) to provide a more challenging GNN evaluation task. 

\begin{table}[h!]
\centering
\begin{tabular}{||c c c c||} 
 \hline
 Trajectory & Re & Snapshots & Category \\ [0.5ex] 
 \hline\hline
 1 & 26214 & 434 & Training \\ 
 2 & 32564 & 386 & Training \\
 3 & 39076 & 396 & Training \\
 4 & 29307 & 378 & Testing \\
 5 & 45589 & 405 & Testing \\ [1ex] 
 \hline
\end{tabular}
\caption{BFS simulation trajectories used to populate datasets for the graph neural network training and evaluation.}
\label{tab:dataset}
\end{table}

Figure~\ref{fig:data_collection} shows the time evolution of the streamwise velocity component at the specified near-step probe locations for all five cases in Table~\ref{tab:dataset}. The velocity signals display highly unsteady decaying periodic behavior indicative of turbulent vortex shedding cycles observed in the BFS configuration at these high Reynolds numbers. These trends ultimately illustrate both the complex dynamics characteristic this configuration, as well as the appreciable difference in dynamics observed in the set of training and testing trajectories. For example, in accordance with the increased Reynolds number, significant increases in velocity magnitude and shedding cycle frequency is seen for Trajectory 5, which is included in the testing set.

\begin{figure}
    \centering
    \includegraphics[width=\columnwidth]{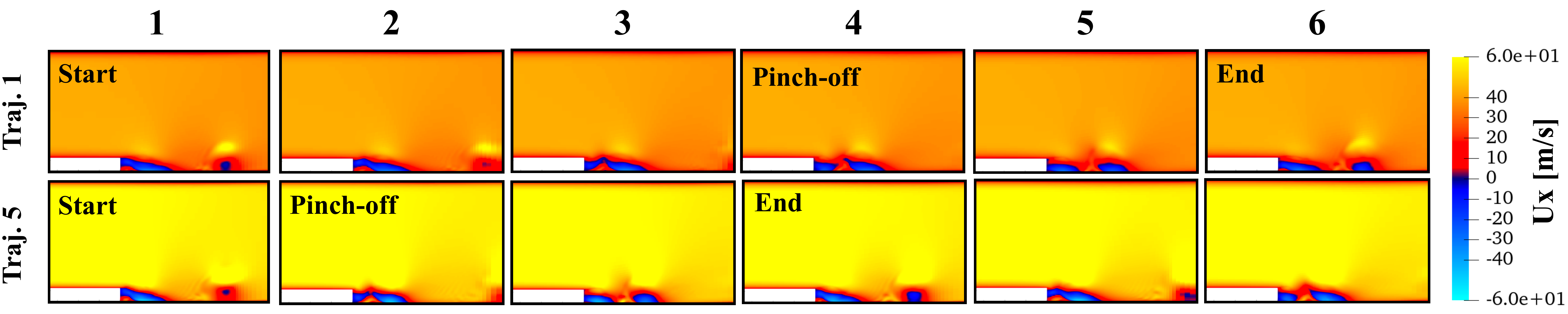}
    \caption{Visualization of BFS shedding cycles via streamwise velocity evolution for trajectory 1 (top, Re=26214) and trajectory 5 (bottom, Re=45589). Columns display instantaneous snapshots separated by a time of $10^{-3}$ seconds.}
    \label{fig:time_evolution_vis}
\end{figure}

These periodic trends in velocity signal are linked to the continual emergence of recirculation zones in the near-step region, causing the re-attachment point for the separated shear layer induced by the step to oscillate along the bottom wall in the x-direction. A qualitative comparison of this re-attachment cycle is shown in Fig.~\ref{fig:time_evolution_vis} for trajectories 1 and 5, which are contained in the training and testing sets respectively. The figure displays a series of instantaneous streamwise velocity snapshots, each separated by a time of $10^{-3}$ seconds, to illustrate the variation in the separation-reattachment process due to the differences in Reynolds number. 

\begin{figure}
    \centering
    \includegraphics[width=0.8\columnwidth]{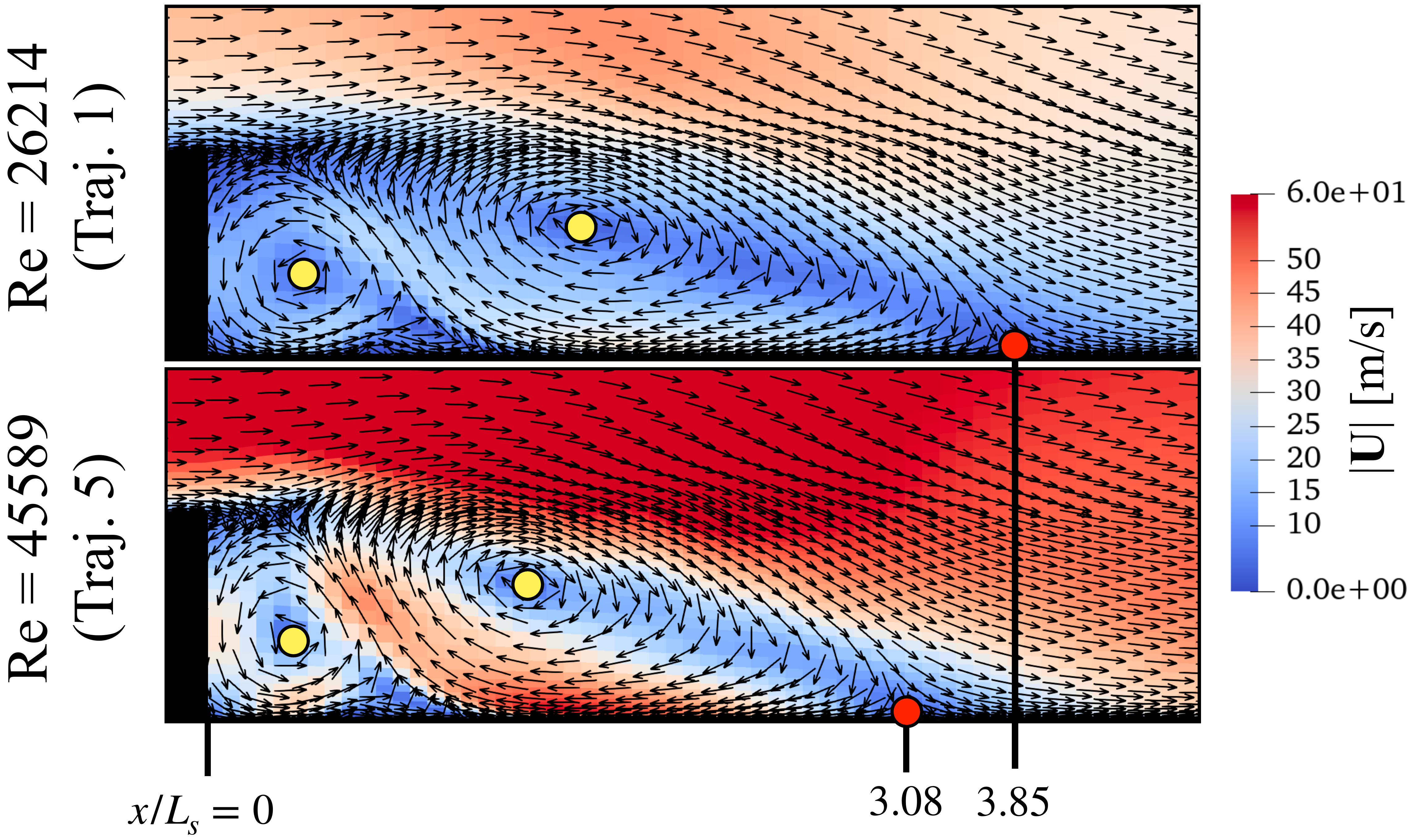}
    \caption{Velocity field comparison near step corresponding to frame 1 (beginning of shedding cycle) in Fig.~\ref{fig:time_evolution_vis} in trajectory 1 (top) and trajectory 5 (bottom). Fields are colored by velocity magnitude, with arrows displaying velocity vector orientation. Yellow circles denote center of recirculation zones, and red circles denote flow reattachment zones.}
    \label{fig:recirculation}
\end{figure}

The snapshot sequences illustrate how the cycle is initiated when the reattachment point experiences a "pinch-off" effect caused by the emergence of a recirculation zone near the step corner, which in turn advects a coherent region of negative streamwise velocity downstream. The pinch-off time in the cycle occurs earlier as Reynolds number increases -- in Fig.~\ref{fig:time_evolution_vis}, this corresponds to snapshot 4 in trajectory 1 versus snapshot 2 in trajectory 5. Additionally, as the pinch-off phenomenon terminates, the cycle concludes when the flowfield encounters a similar condition as the starting point, corresponding to a more uniformly reattached flowfield near the step. As implied through the oscillatory nature of the velocity signals in Fig.~\ref{fig:data_collection}, the visual comparison in Fig.~\ref{fig:time_evolution_vis} shows how the increase in cycle frequency in the higher Reynolds number cases is paired with a proportional decrease in the reattachment length. As shown in Fig.~\ref{fig:recirculation}, drop in reattachment length is identified by stronger recirculation zones near the step caused by increased magnitudes of adverse pressure gradients observed in the flow. 

The above comparisons in training and testing trajectories, along with flowfield visualizations, are provided to give confidence to the fact that the key physical quantities of interest characterizing flow over backward-facing steps (e.g. reattachment, downstream boundary layer development, emergence of recirculation zones in step corner) are correctly captured in the simulation procedure. Although additional effort can be undertaken in the modeling procedure to establish consistency with experimental and direct numerical simulation (DNS) studies at the chosen operating conditions, such analyses are deemed out-of-scope in the context of demonstrating the graph neural network capabilities described in Sec.~\ref{sec:methodology} and \ref{sec:results}.

\section{Methodology}
\label{sec:methodology}
% \begin{itemize}
% Show image of graph neural network architecture 
% List the main components, and say we will describe this below 
% Mention how this is different from the approaches used before: e.g. in u-net, different because we eliminate skips. In multiscale, different because each individual layer is interpreted as a multiscale GNN. 
% Describe graph generation in this part, I think this would work better 
% Then, the rest of the sections discuss: (1) multiscale message passing, (2) top-k pooling, (3) un-pooling 
% \end{itemize}

\begin{figure}
    \centering
    \includegraphics[width=\columnwidth]{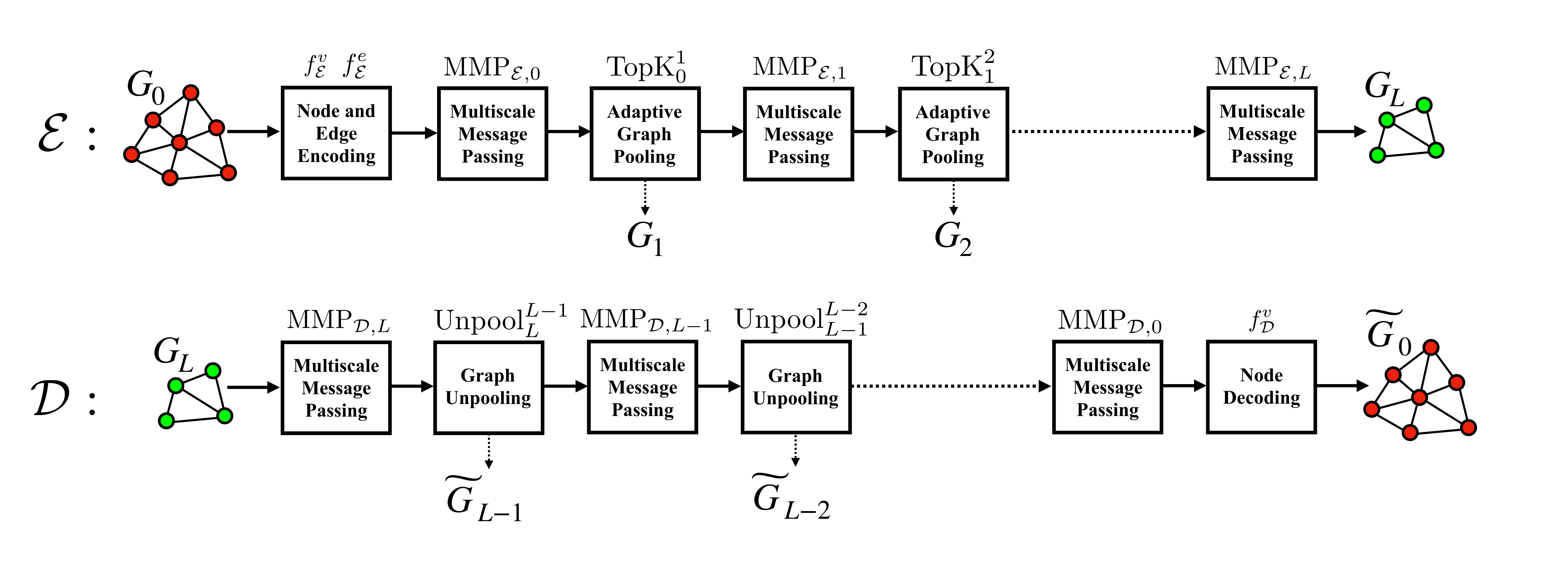}
    \caption{\textbf{(Top)} Encoder flowchart, showing the procedure by which the input graph representation of flowfields $G_0$ (red nodes) is downsampled into the latent graph $G_L$ (green nodes). Main building blocks are multiscale message passing layers (see Sec.~\ref{sec:message_passing}) and Top-K pooling layers (see Sec.~\ref{sec:topk}). \textbf{(Bottom)} Decoder flowchart, showing the procedure by which the latent graph $G_L$ is upsampled to recover $\widetilde{G}_0$, which contains predicted flowfield quantities at the original graph node locations. Upsampling is achieved using an unpooling layer described in Sec.~\ref{sec:topk}.}
    \label{fig:gnn_architecture}
\end{figure}

The graph neural network (GNN) autoencoder architecture is shown in Fig.~\ref{fig:gnn_architecture}. The various layers shown in the schematic are organized into encoder and decoder segments that operate on graph representations of the unstructured flowfield data. The GNN forward pass can expressed concisely through the actions of the encoder and decoder via 
\begin{align}
    \label{eq:autoencoder}
    G_L &= {\cal E} (G_0), \\
    \widetilde{G}_0 &= {\cal D} (G_L),
\end{align}
where ${\cal E}$ and  ${\cal D}$ are the graph-based encoder and decoder, respectively. As shown in Fig.~\ref{fig:gnn_architecture}, the encoder $\cal E$ takes as input a graph $G_0$. The encoder output -- referred to as the  \textit{latent graph} $G_L$ -- is conditioned on both the input graph quantities (e.g. nodes and edge attributes) and the parameters used in the encoder layers. The decoder $\cal D$ takes as input the latent graph $G_L$ and produces the output graph $\widetilde{G}_0$; the decoding evaluation is similarly conditioned on a separate set of layer parameters.

Analogous to conventional autoencoding goals, the objective of the graph autoencoder is (a) to obtain a latent graph $G_L$ that reduces dimensionality of the input graph $G_0$, and (b) to ensure, through an data-based optimization or learning procedure, that the parameters contained in $\theta_{\cal E}$ and $\theta_{\cal D}$ result in an accurate decoding or reconstruction (i.e. $\widetilde{G}_0 \approx G_0$). In this work, the optimization procedure seeks to minimize the mean-squared error (MSE) between only the node attributes in the initial ($G_0$) and decoded ($\widetilde{G}_0$) graphs -- definitions for these node attributes are provided in Sec.~\ref{sec:graph_generation}. An additional requirement central to this work is that of interpretability: the graph autoencoder should be constructed in such a way that the latent space (in this case, the latent graph $G_L$) is physically interpretable. 

% Describe the input graph: graph generation 
To this end, the main building blocks of the GNN architecture are multiscale message passing (MMP) layers and the graph pooling layers. The MMP layer (described in Sec.~\ref{sec:message_passing}) models information exchange throughout the flow domain at multiple lengthscales based on input graph connectivity, analogous to convolution operations on structured grids. The graph pooling layer (detailed in Sec.~\ref{sec:topk}) is the procedure by which graph dimensionality can be reduced via reduction in the number of nodes. Emphasis here is placed on one particular graph pooling strategy known as Top-K pooling \cite{graph_u_nets}, which has been unexplored in recent GNN-based studies for unsteady fluid flows. A primary objective is to show how this Top-K pooling layer, through an adaptive graph reduction procedure that essentially amounts to flowfield-conditioned node sampling and sensor identification, produces interpretable latent graph representations tailored to the regression task at hand.

\subsection{Graph representation of flowfield data}
\label{sec:graph_generation}
Before presenting details of the message passing and pooling operations, the input graph generation procedure must first be described. The input graph $G_0$ is defined as the tuple $G_0 = ({\bf V}_0, {\bf E}_0, {\bf A}_0)$. The quantity ${\bf V}_0 \in \mathbb{R}^{N^v_0 \times F^{v}_0}$ is the input node attribute matrix: a single row contains the $F^v_0$-sized feature vector of one of $N^v_0$ total nodes in the input graph. In the applications considered here, these initial node features represent physical flowfield quantities (i.e. velocity fields sampled at a particular location in physical space), which eventually evolve into hidden node features of a potentially different size upon encountering nonlinear GNN layers during the encoding stage. 

The quantity ${\bf E}_0 \in \mathbb{R}^{N^e_0 \times F^e_0}$ is the input edge attribute matrix: a single row in ${\bf E}_0$ describes the $F^e_0$-sized feature set of a directed edge in the graph. Along with its set of features, a directed edge is characterized by both a sender and receiver node index -- these nodes are deemed "similar" in some sense based on the user-defined criteria used to derive the connection. 

Accumulating these indices for all edges in the graph generates the adjacency matrix ${\bf A}_0 \in \mathbb{Z}^{N^v_0 \times N^v_0}$. The adjacency matrix is used to facilitate arithmetic operations conditioned on the graph connectivity (e.g. the edge aggregation procedure described in Sec.~\ref{sec:message_passing}). Put simply, if the rows of this matrix correspond to sender node indices and columns to receiver node indices, any directed edge can be represented as a value of unity in the corresponding location in the adjacency matrix with zeros everywhere else, such that the number of non-zero entries in ${\bf A}_0$ recovers the total number of directed edges $N^e_0$. Two quantities of interest derived from the adjacency matrix are the node neighborhood and node degree. The neighborhood of a receiver node $i$, denoted $N(i)$, is the set of sender node indices that share an edge; the degree of node $i$, denoted $d(i)$, is the cardinality of this neighborhood set ($d(i) = |N(i)|$). It should be noted that in practice, assuming the average node degree is significantly smaller than the total number of nodes, adjacency matrices are stored in a sparse matrix format to both ensure scalability with respect to $N_v$ and avoid memory limitations. 

\begin{figure}
    \centering
    \includegraphics[width=0.6\columnwidth]{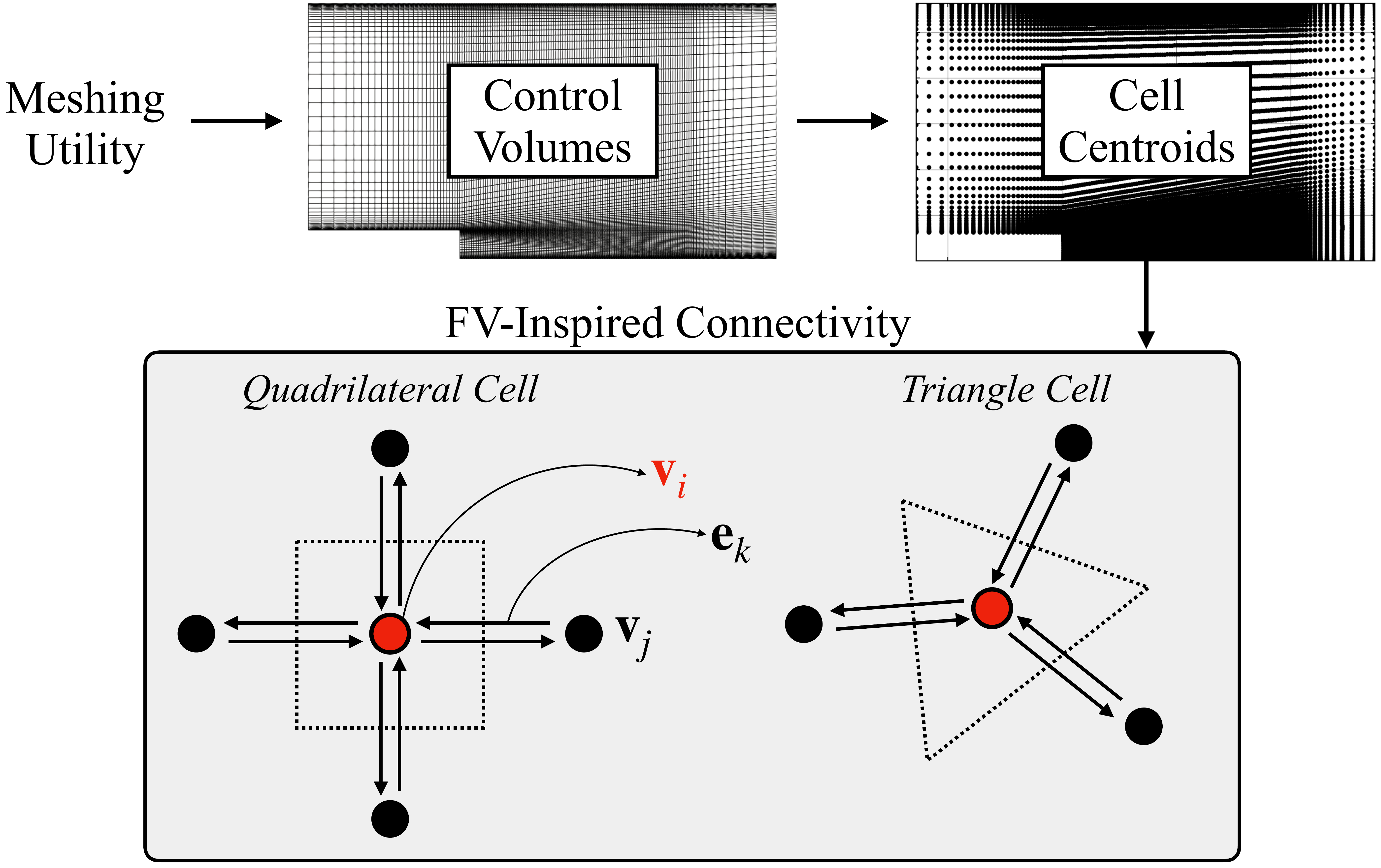}
    \caption{Graph generation procedure used in this work. Graph connectivity is consistent with the finite-volume formulation of CFD numerics. Although only quadrilateral cells are present in the mesh used here, schematics in shaded box show a single neighborhood for two different cell geometries to illustrate the approach. Red circle indicates root node, dashed lines indicate underlying cell, and solid arrows indicate directed edges that intersect cell faces.}
    \label{fig:graph_generation}
\end{figure}

Given the above definitions, the steps required to actually generate the input graph from the flowfield data described in Sec.~\ref{sec:dataset} are (1) produce the set of nodes and corresponding node attributes from the raw data, and (2) construct the list of edges (graph connectivity) using some notion of similarity derived from either all or a subset of these node attributes. To motivate consistency with the finite-volume formulation of numerics utilized within the \verb|OpenFOAM| framework, this work adopts graph connectivity using the latter pathway. As shown in Fig.~\ref{fig:graph_generation}, in this setting, graph nodes represent cell centroids. The node attributes on $G_0$ are then readily initialized as 
\begin{equation}
    {\bf V}_{0,i} = {\bf X}(t_i), \quad i = 1, \ldots, N. 
\end{equation}
where the subscript $i$ is introduced as a snapshot index, and ${\bf X}(t_i)$ is the flowfield snapshot sampled at time $t_i$ as described in Sec.~\ref{sec:dataset}. In the end, the number of nodes in the initial graph $N^v_0$ is equivalent to the number of cells $N_C$ in the mesh (see Fig.~\ref{fig:mesh}). 

Directed edges between nodes are then instantiated based on shared cell faces. As such, edges that connect nodes in this formulation coincide with flux paths between cells. To prevent issues related to mesh skewness and variable mesh density during message passing operations, additional edges are appended to these initial edges by also establishing connections between nodes within a user defined radius of $0.08L_S$. In other words, a radius-based connectivity is superimposed onto the finite volume connectivity, where for a given node, edges are created for all other nodes within the specified length scale. The number of nodes and edges in the initial graph after this procedure are $N_0^v=14476$ and $N_0^e=133739$ respectively. 

In-line with previous approaches, the initial edge feature matrix ${\bf E}_0$ is populated with the distance vector produced by the corresponding sender and receiver node indices. Note that due to the time-dependence of the flowfield ${\bf X}(t)$, the node attributes in the input graph -- the velocity fields at the cell centroids -- are also time-evolving. However, because the physical space coordinates of the mesh cells are fixed, the edge attributes ${\bf E}_0$ and adjacency matrix ${\bf A}_0$ \textit{in the input graph $G_0$} are fixed for all snapshots.

\subsection{Encoding and decoding procedures} 
The GNN evaluation procedure shown in Fig.~\ref{fig:gnn_architecture} is broken down in the subsections below into encoder and decoder components in Sec.~\ref{sec:encoder} and \ref{sec:decoder} respectively. With the general background provided here, the reader is directed to Sec.~\ref{sec:message_passing} and \ref{sec:topk} for finer details on multiscale message passing (MMP) layers and graph pooling layers used within the architecture. 

\subsubsection{Encoder}
\label{sec:encoder}
As implied by the subscript, the input graph $G_0$ resides on the baseline "zeroth" level of the autoencoding architecture. The encoder produces a hierarchy of graphs at higher levels through successive message passing and Top-K pooling operations. The encoding procedure terminates upon encountering the latent graph $G_L$, where $L$ denotes the maximum level in the architecture (the "latent graph" is always referred to here as the graph produced at the highest level). Consistent with the notation introduced in Sec.~\ref{sec:graph_generation}, a graph at the $l$-th level is denoted $G_l = ({\bf V}_l, {\bf E}_l, {\bf A}_l)$, and is characterized by its own set of nodes, edges, and adjacency matrix. For example, in the case of $L=2$, graphs at three different levels are utilized in the architecture: $G_0 = ({\bf V}_0, {\bf E}_0, {\bf A}_0)$ (the input graph at level 0), $G_1 = ({\bf V}_1, {\bf E}_1, {\bf A}_1)$ (the level 1 graph), and $G_2 = ({\bf V}_2, {\bf E}_2, {\bf A}_2)$ (the level 2 or latent graph).

Depending on the reduction factors used in the pooling operations, the goal of the encoder is to ensure that the number of nodes in the graph decreases as the number of levels increases, resulting in the following inequality:
\begin{equation}
    N_0^v > N_1^v > \ldots > N_L^v, 
\end{equation}
where $N_l^v$ ($l = 0, \ldots, L$) denotes the number of nodes at the $l$-the level (equivalent to the number of rows in ${\bf V}_l$). The \textit{global reduction factor} ($RF_G$), an input parameter that characterizes the level of compression achieved in the architecture in terms of node reduction, is $RF_G = N_0^v/N_{L}^v$. As will be seen in the results in Sec.~\ref{sec:results}, there are two important implications of the reduction factor: (1) a higher $RF_G$ is expected produce a more difficult reconstruction task for the decoder, and (2) one can use different graph hierarchies (i.e. different values of the maximum graph level $L$) to arrive at a target $RF_G$. 

\textbf{Embedding node and edge features:}
As demonstrated in Refs.~\cite{battaglia_2018,deepmind_2020}, feature space embeddings significantly improve predictive accuracy in GNN-based models. In this process, shown as the first step in Fig.~\ref{fig:gnn_architecture}, before encountering the message passing and pooling operations required to generate graphs at different levels, the node and edge features in the input graph $G_0$ are modified. This step is carried out as ${\bf V}_0 \leftarrow f^v_{\cal E} ({\bf V}_0)$ and ${\bf E}_0 \leftarrow f^e_{\cal E} ({\bf E}_0)$, where $f^v_{\cal E}$ and $f^e_{\cal E}$ are independently-parametrized multi-layer perceptrons (MLPs) batched over the number of nodes and edges respectively in $G_0$. These MLPs operate in feature space only, and therefore result in modified feature dimensionalities for the nodes and edges in the input graph $G_0$ -- the number of nodes and edges, as well as the graph connectivity, is not changed. As a result, after this feature encoding stage, the node and edge attributes reside in a hidden feature space of fixed size, typically larger than the feature space dimensionality used to initialize the original graph nodes and edges. The resulting node and edge features are referred to as hidden channels. For convenience, the hidden channel dimensionality in this work the same for node and edge attributes after this step ($F_0^v = F_0^e$), and is kept fixed during the forward pass until encountering the final node feature decoding operation to recover the desired reconstructed flowfields.

\textbf{Main encoding layers:}

As alluded to in the beginning of this section, the two layers that serve as the backbone of the GNN-based encoder are a multiscale message passing (MMP) layer and an adaptive graph pooling operation known as Top-K pooling. These are carried out in succession to move from a lower level graph $l$ to a higher level graph $l+1$ with fewer nodes, eventually terminating in the generation of the final latent graph at level $L$. These operations are given by
\begin{subequations}
\label{eq:encoder}
\begin{align}
({\bf V}_l, {\bf E}_l) &\leftarrow \text{MMP}_{{\cal E},l}({\bf V}_l, {\bf E}_l, {\bf A}_l), \quad l = 0, \ldots, L, \\
({\bf V}_{l+1}, {\bf E}_{l+1}, {\bf A}_{l+1}) &\leftarrow \text{TopK}_{l}^{l+1}({\bf V}_l, {\bf E}_l, {\bf A}_l | {\bf p}_l), \quad l = 0, \ldots, L-1, 
\end{align}
\end{subequations}
where $\text{MMP}_{{\cal E},l}$ denotes a multiscale message passing layer present in the encoder $\cal E$ at graph level $l$, and $\text{TopK}_{l}^{l+1}$ denotes the Top-K pooling layer that downsamples graph variables from level $l$ to $l+1$. Note that the MMP layer in Eq.~\ref{eq:encoder}(a) does not modify the graph connectivity ${\bf A}_l$ -- rather, it modifies node and edge features through a neighborhood aggregation procedure described in Sec.~\ref{sec:message_passing}. On the other hand, the pooling operation in Eq.~\ref{eq:encoder}(b) does modify the connectivity through a node sampling procedure conditioned on the feature-wise projection vector ${\bf p}_l$. This sampling mechanism usefully provides interpretability properties in latent graphs -- details on this aspect are provided in Sec.~\ref{sec:topk}. 

\subsubsection{Decoder}
\label{sec:decoder}

The goal of the decoder is to upsample the latent graph $G_L$ back into the original input graph dimensionality in terms of number of nodes and edges. The upsampling procedure is executed with a series of graph unpooling layers that mirror the pooling operations used in the encoding stage. In other words, starting from $G_L$, the successive layer operations encountered in the decoding stage terminate upon arriving at the graph $\widetilde{G}_0$ that shares the same number of nodes and edges as the input graph ${G}_0$. 

\textbf{Main decoding layers:} The decoding procedure is outlined in the following steps:
\begin{subequations}
\label{eq:decoder}
\begin{align}
({\bf V}_l, {\bf E}_l) &\leftarrow \text{MMP}_{{\cal D},l}({\bf V}_l, {\bf E}_l, {\bf A}_l), \quad l = L, \ldots, 0.\\
({\bf V}_{l-1}, {\bf E}_{l-1}, {\bf A}_{l-1}) &\leftarrow \text{unpool}_{l}^{l-1}({\bf V}_l, {\bf E}_l, {\bf A}_l), \quad l = L, \ldots 1. 
\end{align}
\end{subequations}
To mirror the encoding stage, the decoder first utilizes an MMP layer at graph level $l$ in Eq.~\ref{eq:decoder}(a) before encountering the unpooling layer in Eq.~\ref{eq:decoder}(b), which serves to umpsample the node and edge feature matrices to the sizes consistent with those in level $l-1$. The multiscale message passing layer utilizes identical operations as in the encoding stage (Eq.~\ref{eq:encoder}(a)), but contains different parameters (see Sec.~\ref{sec:message_passing}). This combination of message passing and unpooling essentially amounts to a learnable interpolation operation between successive graphs in the Top-K hierarchy -- additional detail on the unpooling procedure is provided in Sec.~\ref{sec:topk}. 

\textbf{Decoding node features:} To recover the final reconstructed flowfield at the nodes, after all message passing and unpooling steps are executed, the decoder utilizes a node-wise feature decoder at level 0 that reverses the feature embedding procedure executed in the first step of the encoding stage. This step is carried out as ${\widetilde{\bf V}_0} \leftarrow f^v_{\cal D}({\bf V}_0)$, where $f^v_{\cal D}$ is an MLP that operates feature-wise, serving to transform the node dimensionality from the hidden channel dimension back into the original dimension corresponding to the number of flowfield observables used in the dataset. The final predicted quantity used to compute the MSE loss function is then $\widetilde{\bf V}_0$, which contains the reconstructed flow quantities (velocity components) at the nodes.

\subsection{Multiscale message passing (MMP) layer}
\label{sec:message_passing}
The notion of message passing was put forward in Ref.~\cite{gilmer_2017} as a general framework to unify a wide range of GNN modeling strategies. In any message passing approach, the goal is to provide a functional representation for the interaction between node attributes in a single neighborhood of the graph. Within a standard message passing layer, this interaction rule is formulated such that (a) its function parameters are shared throughout all neighborhoods in the graph, thereby ensuring a domain-agnostic modeling approach, and (b) it does not modify the input graph connectivity (adjacency matrix). Inspired by methods used in the multigrid community, recent work in GNN-based modeling has demonstrated that augmenting standard message passing operations with a hierarchy of coarse grids naturally results in more effective information propagation \cite{multiscale_meshgraphnet,lino2022multi}. These methods remove the need for modeling large lengthscale interactions with a very large number of message passing operations, which in turn reduces memory requirements during training and drops inference times. 

\begin{figure}
    \centering
    \includegraphics[width=\columnwidth]{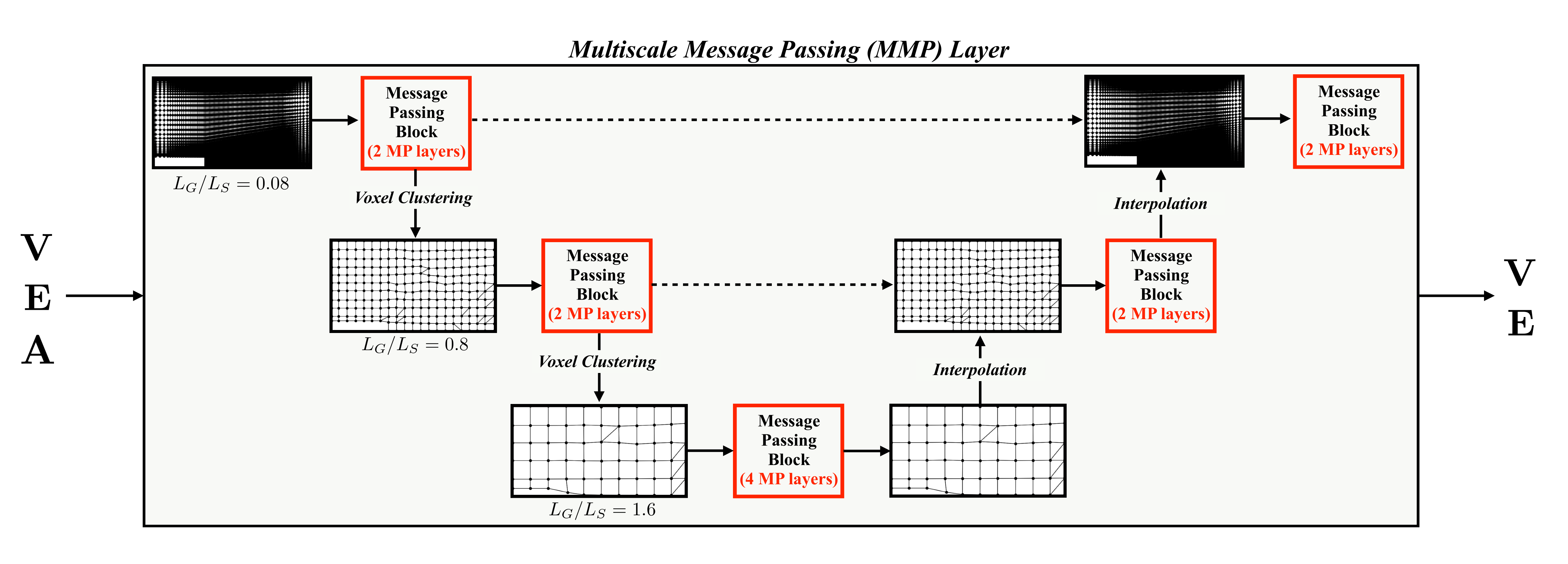}
    \caption{Schematic of MMP layer used in the autoencoding architecture in Fig.~\ref{fig:gnn_architecture}. Message passing blocks (see Fig.~\ref{fig:message_passing}) are shown in red, specifying number of MP layers per at each coarsening level. Message passing lengthscales $L_G$ normalized by step size $L_S$ are specified underneath respective graph images. Dashed lines indicate residual/skip connections for node and edge attributes. Note that these skip connections operate within \textit{a single layer} of either the encoder or decoder (see Fig.~\ref{fig:gnn_architecture}).}
    \label{fig:mmp_layer}
\end{figure}

As the name implies, the multiscale message passing (MMP) layers used in both encoding and decoding stages leverage this principle. A schematic of a single MMP layer is shown in Fig.~\ref{fig:mmp_layer}. The MMP layer consists of a series of message passing blocks, each on a different coarsening level relative to the baseline mesh. Each message passing block contains a fixed number of single-scale message passing (MP) layers. After executing a message passing block, node and edge attributes are interpolated to a coarser graph level characterized by a larger inter-node lengthscale. This proceeds until the coarsest graph is reached, upon which graph quantities are interpolated back upwards to the starting point, utilizing skip connections for node and edge attributes from the downward pass along the way. It is emphasized that the coarsening levels in MMP layers are different from the concept of graph levels introduced earlier, which come from outputs of Top-K pooling operations -- this distinction is made clear in Sec.~\ref{sec:topk}.

The coarsening mechanism used in each MMP layer comes from a voxel-clustering algorithm, which has been demonstrated to work well in previous GNN applications \cite{lino2022multi}. Although other meshing approaches can be utilized (e.g. triangulations \cite{multiscale_meshgraphnet}), the advantage of the voxel clustering approach is that it directly takes a target message passing lengthscale as an input, and does not require an expensive optimization step in the coarsening procedure. In this approach, a graph is coarsened by overlaying a voxel grid in its physical space bounding box. Each cell in this voxel grid is interpreted as a cluster of fixed size corresponding to a target message passing length-scale. Each voxel cell is identified by a parent node, which corresponds to its cell centroid location in physical space. Upon overlaying the voxel grid, underlying fine graph nodes can be assigned to representative voxel cells/clusters via computation of nearest centroids, establishing a parent-child relationship between two successive coarse and fine graphs in the MMP layer. Edges between parent nodes are added if underlying fine graph edges intersect their shared voxel cell face -- the coarse edge attributes are instantiated using the average of attributes belonging to all fine edges that satisfy this intersection. 

The parent-child ownership labels are then used to facilitate interpolation operations required to transfer node attributes between graphs at different coarsening levels. These interpolation procedures follow the methods introduced in Lino et al. \cite{lino2022multi}. In summary, for each coarse and fine graph pair, a new set of edges connecting children nodes to their respective parent node are established, forming an interpolation stencil for node attributes with learnable weights. This stencil is then used to learn coarse-to-fine and fine-to-coarse mappings for node attributes. The reader is pointed to Ref.~\cite{lino2022multi} for more detail on the interpolation method. 

Section~\ref{sec:mp_block} details the operations involved in a single message passing block. Before proceeding, it should be noted that the operations shown in Fig.~\ref{fig:mmp_layer} are equivalent to a graph U-net architecture. The difference here is that this U-net operation corresponds to a \textit{single layer} -- an MMP layer -- operating on one level in the full autoencoding architecture shown in Fig.~\ref{fig:gnn_architecture}. The effect of including multiple coarsening levels in MMP layers in both encoder and decoder stages will be analyzed in Sec.~\ref{sec:results}.

\subsubsection{Message passing block} 
\label{sec:mp_block}

\begin{figure}
    \centering
    \includegraphics[width=0.7\columnwidth]{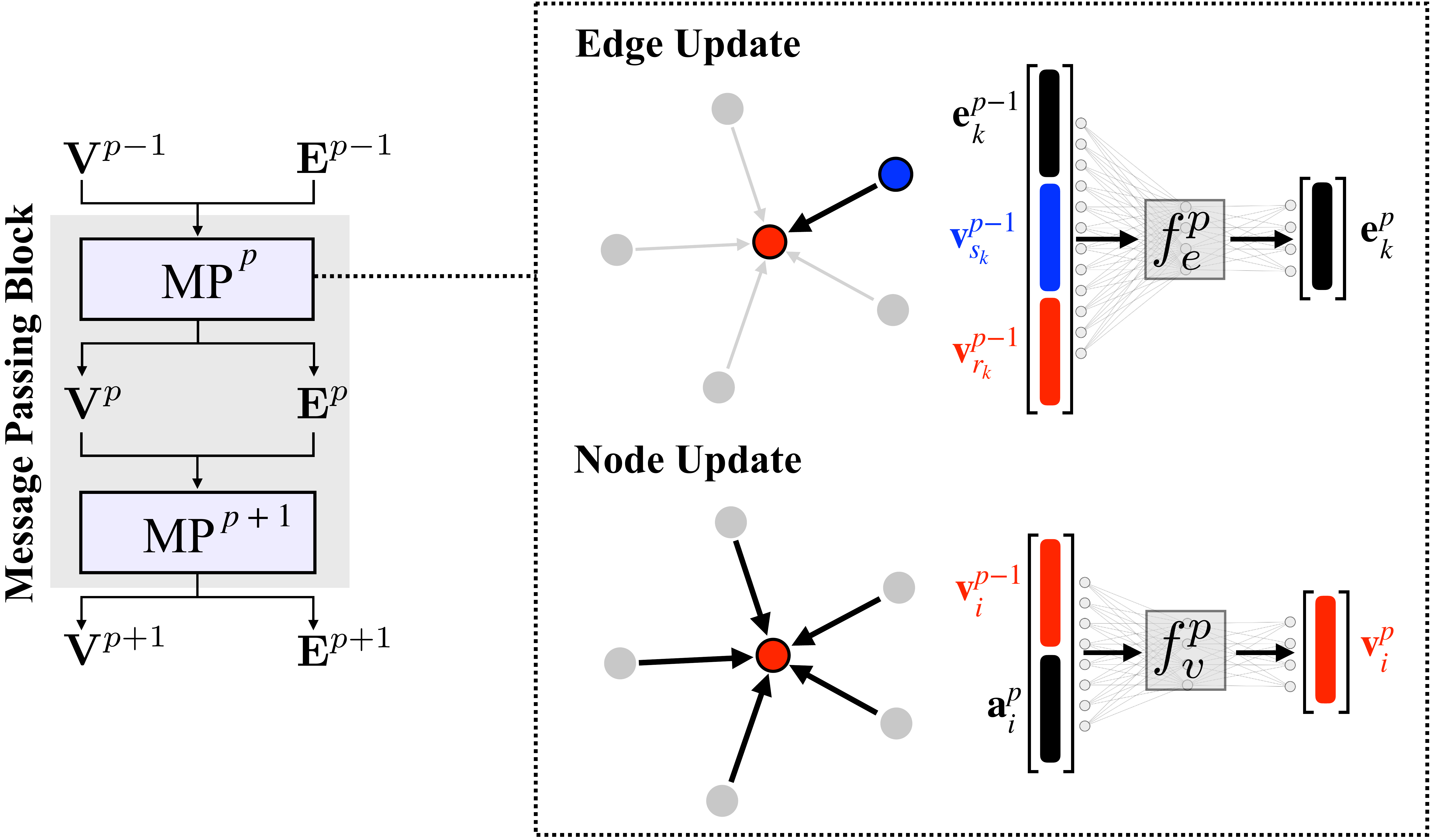}
    \caption{\textbf{(Left)} Schematic of a message passing block consisting of two MP layers. \textbf{(Right)} Illustration of the edge and node update procedures within $\text{MP}^p$ for a single neighborhood. In the edge update (Eq.~\ref{eq:mp}(a)), edge attributes are shown in black, sender node attributes in blue, and receiver node attributes in red. In the node update (Eq.~\ref{eq:mp}(c)), node attributes are shown in red and aggregated edge features in black.}
    \label{fig:message_passing}
\end{figure}

A single message passing block in an MMP layer consists of a fixed number of uniquely parameterized single-scale message passing (MP) layers. The MP strategy used in this work comes from Battaglia et al.~\cite{battaglia_2018}, which has been established in recent years as a well-grounded framework for graph-based modeling of complex physical systems related to fluid dynamics and other applications \cite{deepmind_2020,pfaff2020learning}. The strategy, summarized in Fig.~\ref{fig:message_passing}, is given by the following three steps:

\begin{subequations}
\label{eq:mp}
\begin{align}
\text{Step 1 (Edge Update):} &\quad {\bf e}_k^{p} = {f}_e^p( {\bf e}_k^{p-1} | {\bf v}_{s_k}^{p-1} | {\bf v}_{r_k}^{p-1} ), \quad k = 1, \ldots, n_e, \\
\text{Step 2 (Edge Aggregation):} &\quad {\bf a}_i^{p} = \frac{1}{|N(i)|} \sum_{\{ k:r_k = i \}} {\bf e}_k^{p}, \quad i = 1, \ldots, n_v, \\
\text{Step 3 (Node Update):} &\quad {\bf v}_i^{p} = f_v^p ({\bf v}_i^{p-1} | {\bf a}_i^p), \quad i = 1, \ldots, n_v.
\end{align}
\end{subequations}
Superscripts in the above equations denote the MP layer index local to the message passing block, and vertical bars in function arguments denote concatenation operations. Additionally, ${\bf e}_k^p$ denotes the $k$-th edge attribute vector sourced from the corresponding row in the edge attribute matrix, and ${\bf v}_i^p$ is the $i$-th analogous node attribute vector -- $n_e$ and $n_v$ denote the number of nodes and edges of the graph in question. In Eq.~\ref{eq:mp}(a), the subscripts $s_k$ and $r_k$ present on the right-hand side denote the sender and receiver node indices for the $k$-th edge.

As shown in Fig.~\ref{fig:message_passing}, the functions $f_e^p$ and $f_v^p$ are independent multi-layer perceptrons (MLPs) that act as nonlinear functions producing updated edge features and node features respectively. This allows the update procedures to both capture nonlinear interactions between edge and node attributes from previous layers, and also model complex feedback between individual nodes and their local neighborhoods. Note that the graph adjacency matrix is invoked only in Eq.~\ref{eq:mp}(b) (the aggregation step); the summation is used to transform the edge-based representation of data into a node-based representation via reduction over the edge attributes in the neighborhood $N(i)$ (i.e. ${\bf a}_i^p$ is defined on the nodes).

For some graph $G = ({\bf V}, {\bf E}, {\bf A})$, the MP layer from Eqs.~\ref{eq:mp}(a)-(c) is concisely represented as
\begin{equation}
    \label{eq:mp_concise}
    ({\bf V}^p, {\bf E}^p) \leftarrow \text{MP}^p({\bf V}^{p-1}, {\bf E}^{p-1}, {\bf A}). 
\end{equation}
The $p$-th MP layer, denoted $\text{MP}^p$, updates both node and edge attribute representations without modifying the graph connectivity. Although not shown in Eq.~\ref{eq:mp_concise}, the message passing layer is parameterized by the weights and biases in the MLPs $f_e^p$ and $f_v^p$. As the size of a message passing block increases, the expressive capability of the GNN also increases at the trade-off of computational expense during training and inference stages. Unless otherwise noted, the number of MP layers in each message passing block is provided in Fig.~\ref{fig:mmp_layer}, and parameters are not shared across message passing blocks within any MMP layer.  

\subsection{Adaptive pooling and unpooling layers}
\label{sec:topk}
As shown in Fig.~\ref{fig:gnn_architecture}, the result of an MMP layer operation described in Sec.~\ref{sec:message_passing} is provided to either Top-K pooling layers in the encoding phase, or unpooling layers in the decoding phase. This section provides relevant details on each of these layers.

Analogous to pooling operations in convolutional networks, the primary goal of graph pooling layers is to reduce the degrees-of-freedom of the underlying system being modeled via reduction in the number of nodes. The pooling strategy used in this work -- known as Top-K pooling \cite{graph_u_nets} -- provides a unique take on the reduction process. Instead of achieving reduction via fixed graph coarsening, Top-K pooling layers achieve reduction using \textit{adaptive node sampling}. In other words, the layer reduces the number of nodes by sampling a subset of nodes from the input graph. Node positions in the reduced graph coincide in physical space with nodes in the input graph, but the set of sampled node indices is a function of the input node attributes. As such, if these node attributes are time-evolving, the identified nodes in the reduced graph -- as well as the connectivity -- also change in time. 

This difference in reduction philosophy has led to the Top-K layer being overlooked for the purposes of fluid flow modeling, as the reduced graph does not come from a direct coarsening operation. However, it is shown in this work how the Top-K layer, through the adaptive node sampling procedure, allows for built-in interpretability of the latent graph $G_L$ (see Fig.~\ref{fig:gnn_architecture}). More specifically, the time evolution of latent graphs produced by the pooling operation can be visualized directly in physical space using \textit{masked fields}. Because the node sampling procedure optimizes a regression task, visualization of this latent graph can be used to access regions in physical space relevant to this regression task. It should be noted that this interpretability quality resembles a data-based sparse sensing approach for feature identification, but the advantage here is that this formulation is (a) directly compatible with unstructured grids and complex geometries, and (b) is tailored by design to any user-defined regression problem. In the text below, the high-level details of the Top-K pooling and unpooling operations relevant to this work are described. For additional information on the specifics of Top-K operations, the reader is directed to Ref.~\cite{graph_u_nets}. 

\subsubsection{Top-K pooling and masked fields}

\begin{figure}
    \centering
    \includegraphics[width=\columnwidth]{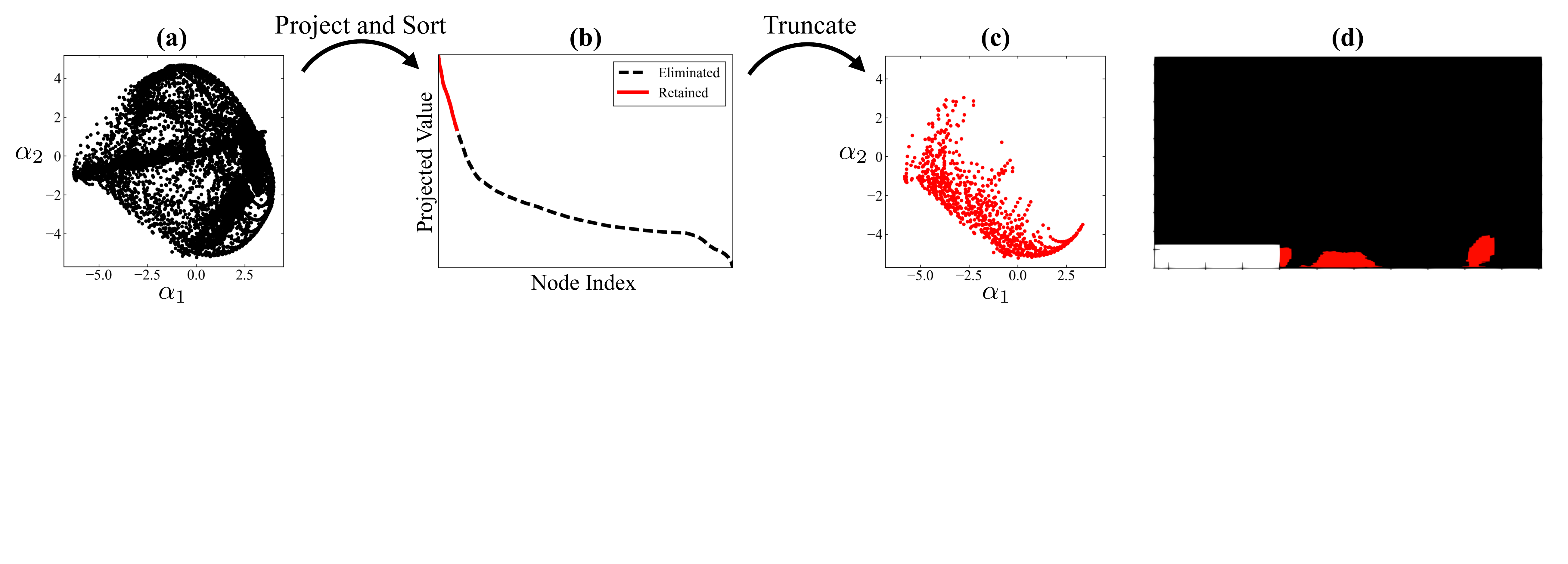}
    \caption{Top-K pooling illustration. \textbf{(a)} Two-component PCA visualization of input node attribute matrix ${\bf V}_0$ associated with an instantaneous flowfield (markers denote individual nodes). \textbf{(b)} 1d projection of ${\bf V}_0$ via Eq.~\ref{eq:topk_projection}. Red region indicates retained node indices corresponding to a reduction factor of 16. \textbf{(c)} PCA visualization of output node attribute matrix ${\bf V}_{1}$ after sampling. Nodes are a subset of those shown in (a). \textbf{(d)} Masked field corresponding to sampled nodes in (c) -- red region indicates where latent graph $G_{1}$ is active.}
    \label{fig:topk_schematic}
\end{figure}

The Top-K pooling operation is given in Eq.~\ref{eq:encoder}(b), and is illustrated in the schematic in Fig.~\ref{fig:topk_schematic}. Two parameters are required: a learnable projection vector ${\bf p}_l$, and the number of nodes $K$ to retain in level $l+1$. In practice, $K$ is recovered from a local reduction factor $RF_l$ via $K = N^v_{l}/RF_l$, where $N^v_{l}$ is the number of nodes at input level $l$. In a first step, the projection vector ${\bf p}_l \in \mathbb{R}^{F_l^v \times 1}$ is used to transform nodes residing in the input $F_l^v$-dimensional feature space into a one-dimensional representation. In a second step, the projected node features are ranked in descending order and truncated such that the indices of the topmost $K$ nodes are retained. These $K$ node indices become "sampled" nodes, forming the node attribute vector ${\bf V}_{l+1}$ on the reduced graph (i.e. $K = N^v_{l+1}$).

More formally, the initial projection step is given by 
\begin{equation}
    \label{eq:topk_projection}
    {\bf y}_l = \frac{ {\bf V}_{l} {\bf p}_l }{ \lVert {\bf p}_l \rVert } \in \mathbb{R}^{N^v_l \times 1}, 
\end{equation}
where ${\bf y}_l$ is the projected vector. The node selection step is given by 
\begin{equation}
    \label{eq:topk_truncation}
    {\cal I}_{l+1} = \text{rank}({\bf y}_l, K), 
\end{equation}
where ${\cal I}_{l+1}$ is the set containing the node indices of the graph at level $l$ corresponding to the topmost $K$ quantities of the sorted projections in ${\bf y}_l$. With ${\cal I}_{l+1}$, an indexing operation can be carried out to recover the reduced graph quantities ${\bf V}_{l+1}$, ${\bf E}_{l+1}$, and ${\bf A}_{l+1}$ in a sampling stage, thereby terminating the Top-K pooling layer. It should be noted that in order to ensure the projection vector ${\bf p}_l$ can be trained using backpropagation, a gating function in the form of a sigmoid activation is performed on the downsampled node features. For details on this gating procedure, see Ref.~\cite{graph_u_nets}. 

\begin{figure}
    \centering
    \includegraphics[width=0.3\columnwidth]{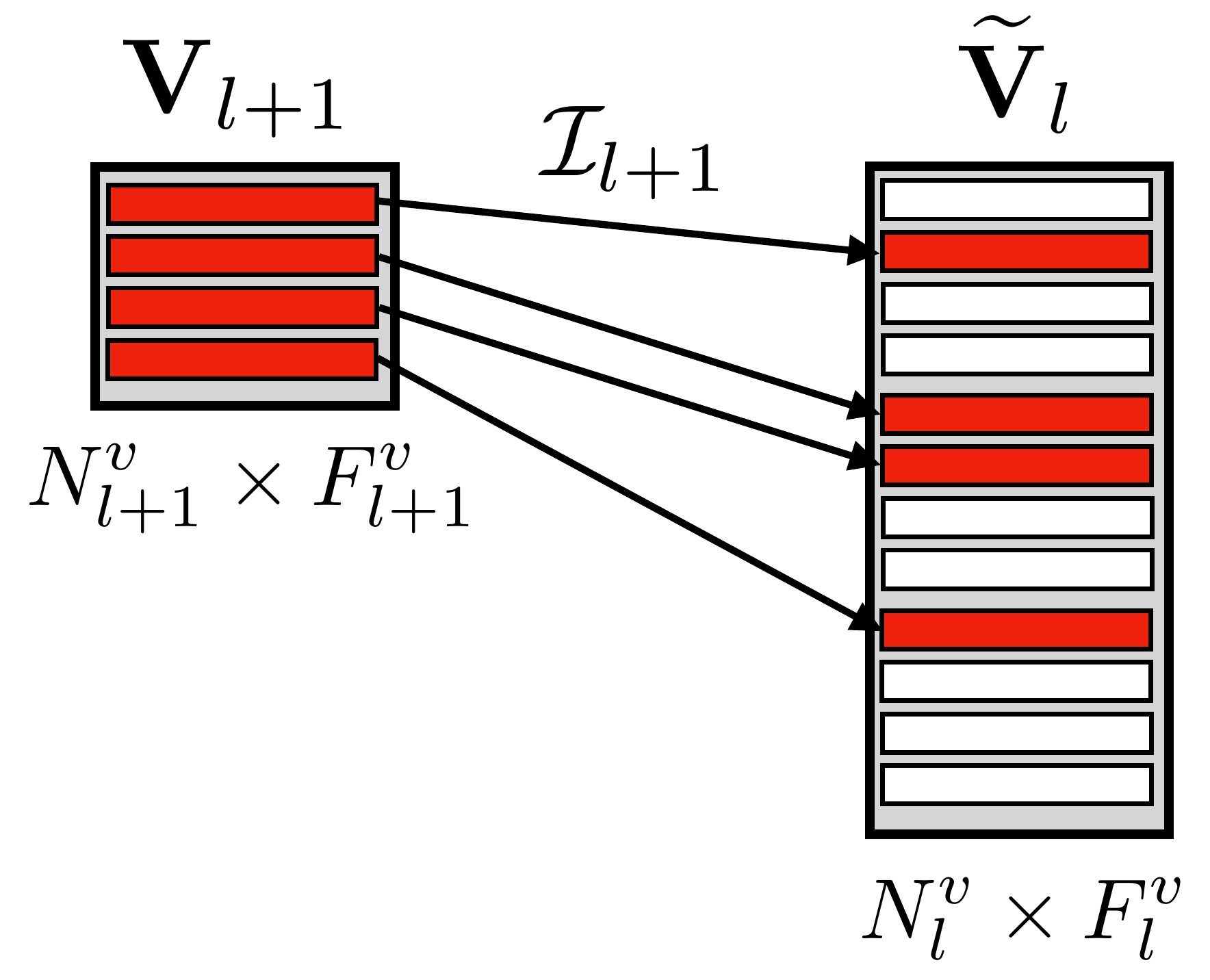}
    \caption{Illustration of the distribution operation (Eq.~\ref{eq:distribute}) used in unpooling. Red boxes denote interpolated values and white boxes denote zeros.}
    \label{fig:unpooling}
\end{figure}

\textbf{Interpretable masked fields:} Crucial to the GNN architecture in Fig.~\ref{fig:gnn_architecture} is that the Top-K pooling operation is used to construct a hierarchy of \textit{adaptive} graphs (i.e. non-fixed adjacency matrices) -- the node positions in physical space of the input graph $G_0$ are fixed in time, but the node positions for levels $l > 0$ change in time. This is because, for a fixed projection vector ${\bf p}_l$ obtained after training, indices of the sampled nodes at level $l+1$ adapt to the flow features contained in the input node attribute matrix ${\bf V}_0$, which are time-evolving. Since each of these indices corresponds to a particular location in physical space, the evolution of reduced graphs at levels $l > 0$, as well as the final latent graph at level $L$, can be visualized via \textit{masked fields}. The masked field, an example of which is shown in Fig.~\ref{fig:topk_schematic}, is a visualization of the level index $l$ of each cell in the physical domain. Utilized further in Sec.~\ref{sec:results}, these masked fields reveal the GNN's ability to (a) physically interpret latent graphs, and (b) identify evolving coherent structures in the domain intrinsically tied to the regression task (flowfield reconstruction).

\subsubsection{Unpooling}
To facilitate graph decoding, the unpooling operation in Eq.~\ref{eq:decoder}(b) can be performed in the Top-K framework so long as ${\cal I}_{l+1}$ is available. As illustrated in Fig.~\ref{fig:unpooling}, the unpooling layer is interpreted as a data distribution operation that resembles delta-function interpolation. The distribution operation for node features is given by 
\begin{equation}
    \label{eq:distribute}
    \widetilde{\bf V}_{l} = \text{distribute}({\bf V}_{l+1}, {\cal I}_{l+1}, {\bf 0}_l). 
\end{equation}

In the above equation, the matrix ${\bf 0}_l \in \mathbb{R}^{N^v_l \times F^v_l}$ contains only zeros -- it is interpreted as an empty node matrix existing on level $l$. The unpool operation "fills in" (or replaces) rows of the matrix ${\bf 0}$ with the corresponding rows of ${\bf V}_{l+1}$ as per the indices contained in ${\cal I}_{l+1}$. Edge attribute matrices $\widetilde{\bf E}_{l}$ can be recovered in an analogous fashion, leading to the full unpooling operation of Eq.~\ref{eq:decoder}(b). Although the operation in Eq.~\ref{eq:distribute} successfully recovers the original data dimensionality at level $l$ (which is the primary goal of the unpooling layer), since skip connections are not used, many values in the upsampling node attribute matrix ${\bf V}_{l}$ can be zero depending on the level of reduction (i.e. value of $K$). The role of the MMP layer after unpooling (see Fig.~\ref{fig:gnn_architecture}) is to treat this issue by efficiently distributing values in the unpooled graph to fill in empty nodes.

\section{Results}
\label{sec:results}

In this section, analysis of the GNN-based autoencoder is performed for the flowfield reconstruction task from a variety of angles, with the primary scope aimed at showcasing the interpretability and reconstruction properties of generated latent spaces (or latent graphs). For all models discussed herein, the objective function used during training is the mean-squared error between input and reconstructed node feature matrices. Three-layer MLPs were used for node/edge attribute encoding and decoding, and two-layer MLPs were used for node and edge updaters in message passing operations. In all cases, MLPs utilize exponential linear unit (ELU) activation functions \cite{elu} with residual connections. Layer normalization \cite{layernorm} is applied after each MLP evaluation. Input and reconstructed node features consist of streamwise ($u_x$) and vertical ($u_y$) velocity components that are standardized using training data statistics. 

GNN architecture development and training was performed using a combination of PyTorch \cite{pytorch} and PyTorch Geometric \cite{pytorch_geom} modules. For training, flowfield snapshots obtained from trajectories 1-3 (see Table~\ref{tab:dataset}) were used, with a randomly selected 10\% of these snapshots set aside as the validation set. During training, a batch size of $8$ was used as per the graph-based mini-batching scheme of Ref.~\cite{pytorch_geom}, and the Adam optimizer \cite{adam} with a learning rate scheduler based on validation loss thresholding was applied. All models were trained using a single Nvidia A100 GPU housed on a node on the Polaris high-performance computer at the Argonne Leadership Computing Facility.

The section proceeds as follows. In Sec.~\ref{sec:coarsening}, the effect of including coarsening operations in the MMP layer on the latent graph and reconstructed fields is discussed. In Sec.~\ref{sec:topk_levels}, for a fixed reduction factor, analysis of changes in the masked fields due to variation in the number of Top-K levels $L$ is conducted. Lastly, in Sec.~\ref{sec:reduction}, the effects of both the node reduction factor and the number of hidden node channels used to characterize the latent graph are studied from the perspectives of reconstruction accuracy and masked field evolution. 

\subsection{Effect of coarsening in the MMP layer} 
\label{sec:coarsening}

\begin{figure}
    \centering
    \includegraphics[width=\columnwidth]{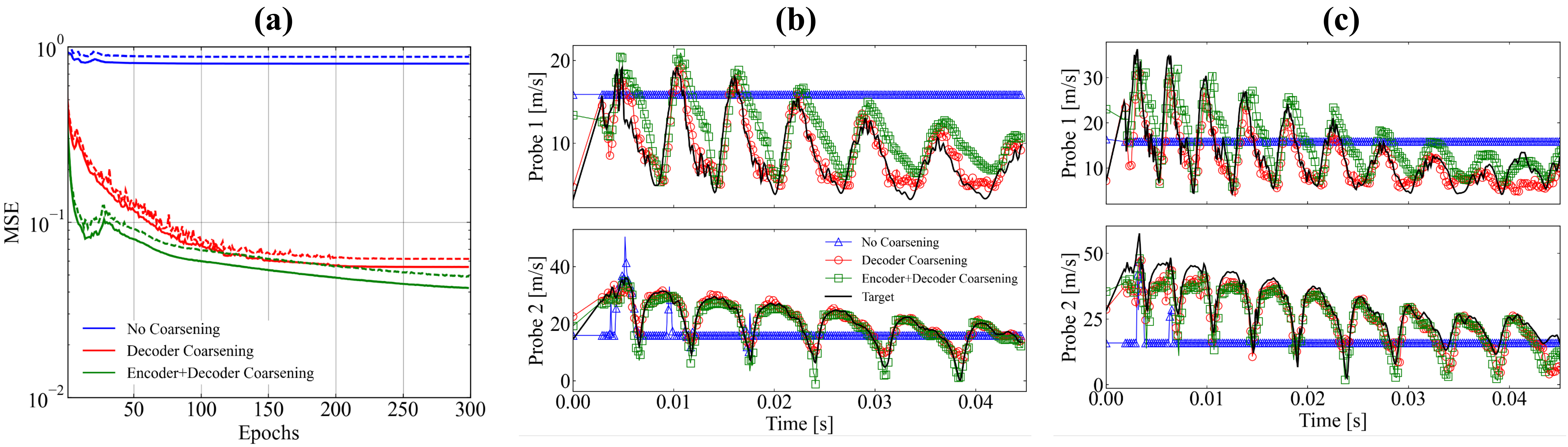}
    \caption{\textbf{(a)} MSE versus training epochs for Model 1 (blue curves), 2 (red curves) and 3 (green curves). Solid lines are training set evaluations, dashed lines are validation set. \textbf{(b)} Reconstructions of streamwise velocity ($u_x$) for Trajectory 4 (Re=29,307, see Table~\ref{tab:dataset}) at probe locations given in Fig.~\ref{fig:data_collection}. Top plot is Probe 1, bottom plot is Probe 2. \textbf{(c)} Same as (b), but for Trajectory 5 (Re=45,589).}
    \label{fig:loss_function_probe}
\end{figure}

To convey the impact of the coarsening operations present in the MMP layers (see Fig.~\ref{fig:mmp_layer}), Fig.~\ref{fig:loss_function_probe} presents mean-squared error histories and streamwise velocity field reconstructions at the probe locations for three different GNN architecture configurations: 
\begin{itemize}
    \item \textbf{Model 1 -- No Coarsening:} In this setting, the coarsening operations in the MMP layers are not used. More specifically, the MMP layer architecture shown in Fig.~\ref{fig:mmp_layer} is modified to neglect all clustering operations used to coarsen the input graph -- as such, message passing proceeds in the conventional fashion without coarsening. This effectively turns the MMP layer into a single-scale message passing block. 
    \item \textbf{Model 2 -- Decoder Coarsening:} In this setting, coarsening operations present in the MMP layers in Fig.~\ref{fig:mmp_layer} are used \textit{only in the decoder} $\cal D$. In the encoder $\cal E$, no coarsening operations are used. 
    \item \textbf{Model 3 -- Encoder+Decoder Coarsening:} In this setting, the MMP coarsening operations are used in both encoder and decoder phases. 
\end{itemize}
When coarsening is used (Models 2 and 3), the MMP layer is consistent with Fig.~\ref{fig:mmp_layer}. When coarsening is not used, all coarse grids corresponding to higher lengthscales in Fig.~\ref{fig:mmp_layer} are ignored, rendering the MMP layer equivalent to a total of 4 standard single-scale message passing layers. 

All three configurations utilize a maximum level of $L=1$ (only one Top-K layer is used), a node reduction factor of $RF_G = 16$ (the number of nodes in the latent graph is dropped by 16x), and a hidden channel dimensionality of $32$. Note that despite the fact that the number of nodes has decreased, the latent graphs here do not provide true compression due to the increase in node feature dimensionality. Assessment of true compression is delayed to Sec.~\ref{sec:reduction}.  Using the above three configurations, the focus of this section is to instead demonstrate the need for coarsening operations in the MMP layer from the perspectives of reconstruction accuracy and masked field interpretability. 

Since the optimization of parameters in the message passing layers and Top-K layers are coupled, the way in which nodes are subsampled to produce the latent graphs is expected to be dependent on the message passing scheme. To assess this effect quantitatively, training history for each of the above three model configurations is shown in Fig.~\ref{fig:loss_function_probe}(a). The impact of coarsening is evident in the loss function trends, as the converged errors for Model 1 (the configuration without coarsening in the MMP layer) shown in the blue curves in Fig.~\ref{fig:loss_function_probe} is an order of magnitude higher than the configurations that include coarsening operations (Models 2 and 3). Further, the jump in accuracy provided by appending additional coarsening operations in the encoding stage (Model 2 to 3) is much lower than when appending coarsening operations in only the decoding stage (Model 1 to 2), implying that the contribution of reconstruction accuracy comes primarily from latent graphs produced using only decoder coarsening operations. 

Figures~\ref{fig:loss_function_probe}(b) and (c) show how the convergence trends during training translate to streamwise velocity reconstructions on the unseen testing trajectories. Prediction trends are ultimately consistent with the loss function curves, in that Model 1 completely fails to capture dynamical information content in the near-step region at all tested Reynolds numbers. On the other hand, reconstructions produced by models that include coarsening are much more in-line with ground truth velocities -- interestingly, Model 2 reflects more accurate flowfield reconstructions than Model 3 near the step, despite the fact that it utilizes coarsening operations only in the decoding stage. As will be seen in the qualitative comparisons below, the improved reconstructions provided by Model 3 come into play in the freestream regions, which is not reflected in the probe measurements shown in Fig.~\ref{fig:loss_function_probe}. Overall, these trends show how (a) coarsening operations are necessary to produce latent graphs that reliably reconstruct the flowfield, and (b) acceptable near-step velocity reconstructions at unseen Reynolds numbers, included extrapolated Reynolds numbers, are produced by autoencoders utilizing multiscale message passing.

\begin{figure}
    \centering
    \includegraphics[width=\columnwidth]{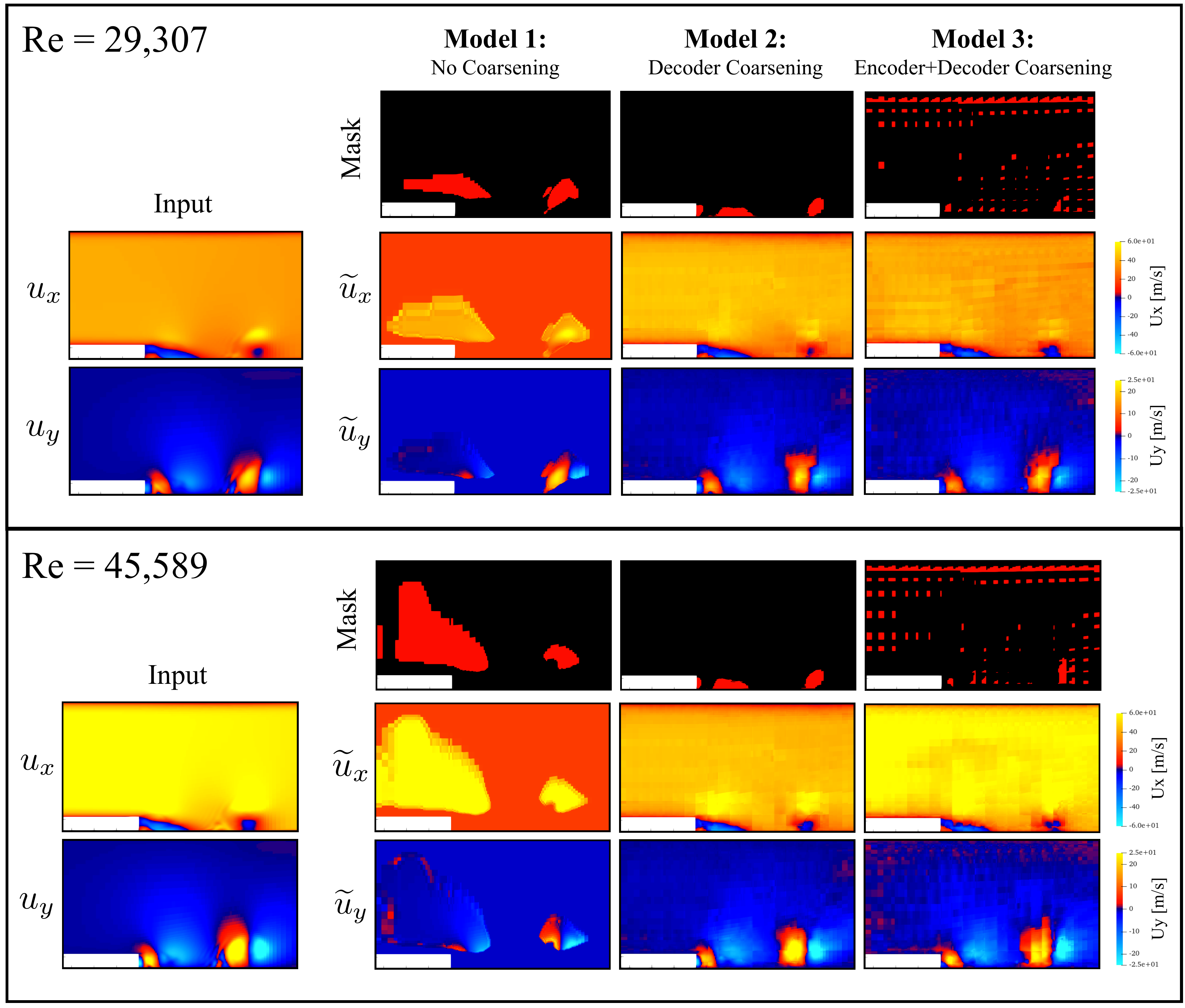}
    \caption{\textbf{(Top)} Autoencoding procedure for an instantaneous flowfield from Trajectory 4 (Re=29,307) using Models 1-3. Leftmost column shows input velocity fields. Corresponding rows show reconstructed fields. Topmost rows show masked fields, where cells colored in red indicate active nodes in latent graph. \textbf{(Bottom)} Same as top, but for a Trajectory 5 snapshot (Re=45,589).}
    \label{fig:combined_prediction}
\end{figure}

The advantage of this autoencoding approach is that the physical characteristics of the reconstruction procedure, as well as the impact of including coarsening operations during message passing via the MMP layers, can be directly interpreted through a visualization of the identified latent graphs ($G_1$ in this case) via masked fields. As such, to supplement the discussion surrounding Fig.~\ref{fig:loss_function_probe}, visualizations of the autoencoding procedure for a single input velocity field snapshot are shown in Fig.~\ref{fig:combined_prediction} for the same three model configurations. For each of the testing set trajectories (an interpolated and extrapolated Reynolds number), the input snapshot shown in Fig.~\ref{fig:combined_prediction} corresponds to step 1 of the reattachment cycle described in Fig.~\ref{fig:time_evolution_vis}.

The masked fields, which indicate where in physical space the latent graphs are active, allow one to visualize locations in the domain corresponding to optimal flowfield reconstruction for the respective models. Note that because the cell sizes vary with spatial location, a fixed node reduction factor (16 for the models shown in Fig.~\ref{fig:combined_prediction}) can result in different identified "volumes" in the masked fields depending the learned Top-K projection vector -- for example, although the number of latent graph nodes is the same, the masked regions identified by Model 2 are smaller in physical space than those in Model 1 because the cell resolution in the identified region is accordingly smaller. Upon visual inspection, a common ground in all three models is that the masked fields in Fig.~\ref{fig:combined_prediction} are characterized by disjoint clusters of identified nodes (red regions), which is in-line with the physical nature of the reattachment cycle described in the discussion surrounding Fig.~\ref{fig:time_evolution_vis}. However, aside from this quality, the masked fields are markedly different for each of the model configurations, which is expected due to the different utilization of coarse grids during message passing. 

The latent graphs identified by Model 1 (no coarsening), for example, identify larger coherent regions in the freestream region above the step than the other two configurations. This is in line with expectations in a single-scale message passing model: since the message passing scheme without coarsening is less efficient at distributing information from the masked region to the rest of the domain, it is understandable that the identified nodes in the latent graph occupy a larger coherent region in space. The disadvantage of the Model 1 configuration is evident when looking at the reconstructed flowfields, which directly show how the lack of multi-scale message passing results in a failure of the model to reconstruct the flow in regions away from the identified mask. Although adding additional message passing layers in a single-scale setting can address this issue, the computational costs and memory limitations during backpropagation render such approaches infeasible. 

In contrast, the reconstructions produced by Models 2 and 3 are successful in recovering the primary features of the BFS flowfield, such as the re-attachment point and the propagating recirculation zone downstream -- although reconstruction accuracy is still imperfect and there are lingering non-physical artifacts in the freestream, Fig.~\ref{fig:combined_prediction} shows how adding coarsening operations via MMP layers addresses the challenging task of recovering full flowfield information from the masked region. 

Although both Models 2 and 3 are able to efficiently propagate information from the masked field, the masked fields themselves are fundamentally different. Model 2 (decoder coarsening only) recovers a masked region that identifies coherent structures consistent with dynamically active regions in the BFS configuration (shear layers, recirculation zones, downstream shedding). On the other hand, Model 3 shows how the inclusion of coarsening operations in the encoder MMP layers drastically alters the coherency of the masked field. Interestingly, the masked field in Model 3 itself resembles a coarse grid unlike the much more spatially coherent Model 2 counterpart. This implies that the inclusion of coarsening operations in the encoding stage effectively trades physical space coherency and interpretability for improved reconstruction accuracy. This is evidenced in the reconstructed fields at the extrapolated Reynolds number (bottom part of Fig.~\ref{fig:combined_prediction}), for which Model 2 fails to reliably extrapolate to unseen freestream velocities when Model 3 succeeds. Despite this, even at extrapolated Reynolds numbers, Model 2 is still able to showcase the powerful ability to reasonably capture the primary BFS flow features while retaining the quality of coherent structure identification in the masked field. Additionally, it should be noted that across all three model configurations, the masked fields retain similar structure at the different Reynolds numbers shown in Fig.~\ref{fig:combined_prediction}, which serves as a form of validation for the identified regions. The regions identified by Model 2 (decoder coarsening) in particular are almost identical at different Reynolds numbers for the snapshots shown. 

The objective of this section was to demonstrate the need for coarsening operations in message passing layers in the decoding task, while also showcasing the primary interpretability property of latent graphs produced by the GNN autoencoder. The architectures utilized above were limited to single levels ($L=1$). As alluded to in Sec.~\ref{sec:methodology}, a beneficial property of this autoencoder is its ability to create a hierarchy of Top-K levels ($L>1$), which in turn leads to alternative representations of the masked fields and enables higher levels of flowfield compression. These aspects are discussed in the sections below. For the remaining analysis in the subsections below, all models configurations correspond to the Model 2 setting (decoder coarsening only) in light of its improved masked field coherency.

\subsection{Number of Top-K levels}
\label{sec:topk_levels}

\begin{figure}
    \centering
    \includegraphics[width=\columnwidth]{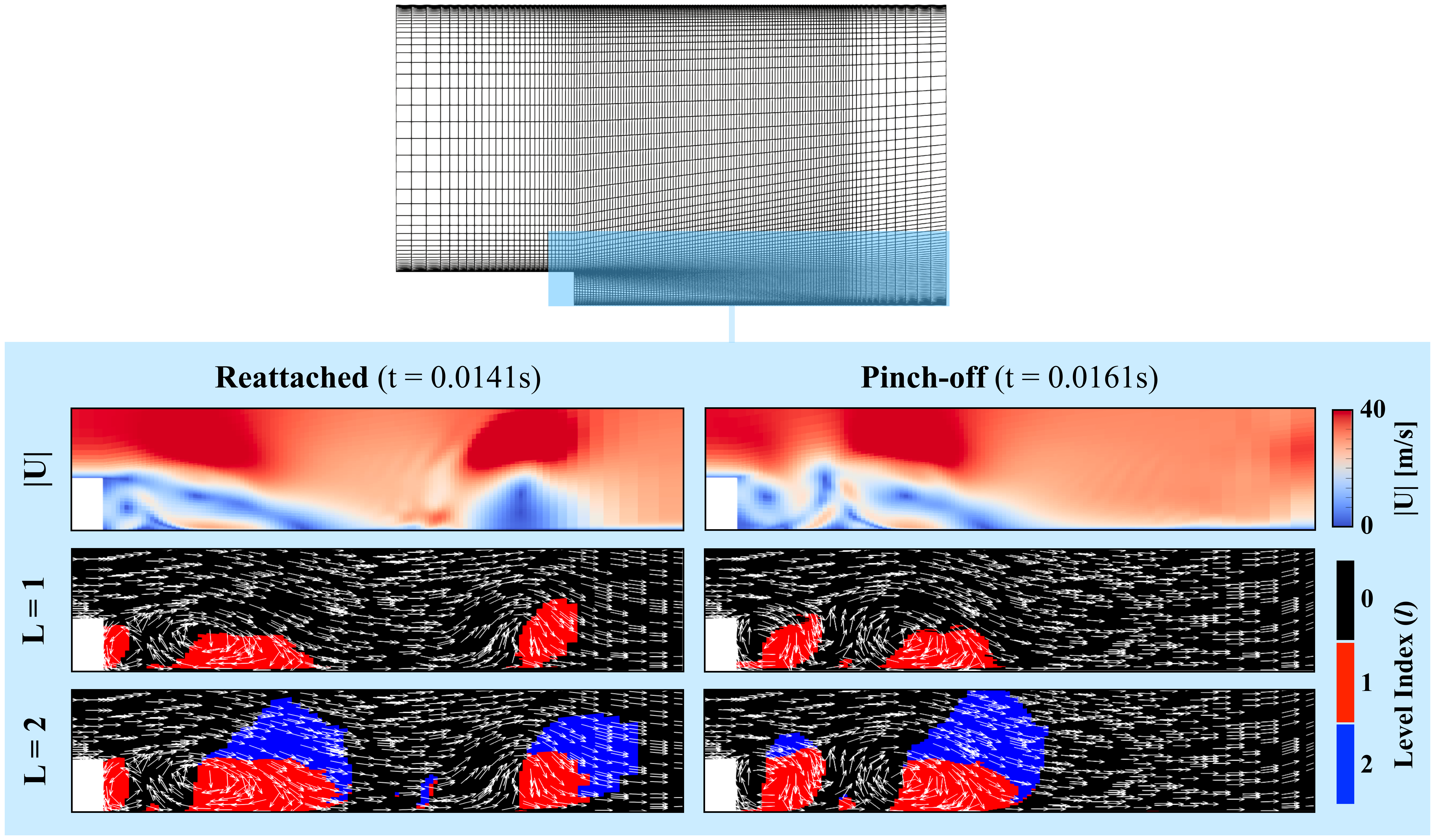}
    \caption{\textbf{(Left)} From top-to-bottom: velocity magnitude field, masked field for $L=1$ model, and masked field for $L=2$ model. Snapshot depicts reattached state in Trajectory 4 (Re = 
29,307) \textbf{(Right)} Same as left, but for a different snapshot in in Trajectory 4 corresponding to the pinch-off phenomenon (see Fig.~\ref{fig:time_evolution_vis}).}
    \label{fig:num_levels}
\end{figure}

The connectivity properties of the latent graph $G_L$ produced by the encoder are not only dependent on the input flowfield, but also on the maximum level $L$ used in the Top-K hierarchy. In other words, for a target global node reduction factor $RF_G$, one can use either a single Top-K level to achieve this reduction ($L=1$, which was used in Sec.~\ref{sec:coarsening}), or a series of Top-K levels, each with smaller local reduction factors $RF_l$. To this end, to isolate the effect of the maximum Top-K level $L$ on the latent graph, Fig.~\ref{fig:num_levels} compares the previously discussed Model 2 from Sec.~\ref{sec:coarsening} for which one Top-K pooling operation was used ($L=1$), to a two-level ($L=2$) counterpart. The $L=2$ model achieves the target global reduction factor of $RF_G=16$ by accumulating two successive reductions that each drop the number of nodes by a factor of 4 ($RF_l=4$). Note that the $L=2$ counterpart here still utilizes the Model 2 configuration described in Sec.~\ref{sec:coarsening}, in that coarsening operations in the MMP layers are invoked only in the decoder.

To better illustrate the changes to the latent graph structure due to time evolution, shown in Fig.~\ref{fig:num_levels} are masked fields for two snapshots sourced from Trajectory 4, each at different stages in the BFS shedding cycle. Additionally, for ease of visualization, masked field plots are overlaid with velocity field orientation vectors to facilitate correlation of the identified coherent structures with recircualtion zones and other flowfield patterns. 

In Fig.~\ref{fig:num_levels}, for the $L=1$ case, the masked field depicts a single identified sub-graph (red region) consisting of 16x fewer nodes than the baseline graph. On the other hand, in the $L=2$ case, the model identifies a hierarchy of subgraphs in accordance with the Top-K procedure: the highest-level latent graph at $l=2$, denoted by the blue regions in the corresponding masked field, also contains (a) 16x fewer nodes than the baseline graph (black regions), and (b) 4x fewer nodes than the identified graph at level 1 (red region). In other words, the masked field, through visualization of the identified level indices in Top-K pooling procedure, allows the user to interpret directly the subsampled hierarchy until the latent graph is reached (highest level index). Note that in the $L=2$ case, the identified latent graph in blue is not disjoint from the intermediary graph at $l=1$, but rather is a subset of the identified nodes at the preceding level.

Figure~\ref{fig:num_levels} shows how changes to the Top-K hierarchy modifies the identified latent graph structure, despite the fact that the number of nodes comprised in these latent graphs is the same. For example, in the $L=2$ case, adding an additional level to the hierarchy allows the latent graph to occupy different regions in physical space that are slightly above the step cavity when compared to the $L=1$ counterpart. This comes directly from the enabled exchange of information between successive levels during the autoencoding procedure when when more Top-K levels are used: in the $L=2$ case, the identified graph at the first level ($l=1$, red region) occupies a larger portion of the domain, enabling the latent graph to target regions of interest unreachable by the single-level model.

Interestingly, the subsampled nodes in the latent graph for the $L=2$ case do not focus on the recirculation zone immediately near the vertical step wall -- instead, focus is placed on downstream regions associated with vortex shedding and the reattached shear layer. When considering all levels together, however, qualitatively similar coherent structures are identified regardless of the parameter $L$ used, giving confidence to the physical significance of the latent graphs at a target global reduction factor. 

When assessing the velocity field patterns in combination with the masked fields in Fig.~\ref{fig:num_levels}, it is evident that the masked fields are correlated with recirculation zones in the flow. For example, in the reattached state, three disjoint structures are visible in the masked field in accordance with the presence of three reciculation zones: a weaker zone near the vertical step wall, another near the reattachment point, and a third corresponding to a stronger shedding vortex approaching the outflow. Of these three regions, the latent graph in the $L=2$ model places more importance on the latter two, implying that the recirculation zones away from the step are more crucial to the flowfield reconstruction task. This is consistent even at different time instances in the unsteady flowfield: for example, in the snapshot exhibiting the pinch-off phenomenon that instantiates the downstream shedding procedure (Fig.~\ref{fig:num_levels}, right), two recirculation zones are present -- as such, two large-scale coherent structures are identified in both masked fields, with the $L=2$ model again emphasizing regions slightly further away from the cavity.

\subsection{Analysis of reduction factor} 
\label{sec:reduction}

The above sections showcased the architectural effects, in terms of coarsening operations utilized in the MMP layers (Sec.~\ref{sec:coarsening}) and Top-K hierarchy size (Sec.~\ref{sec:topk_levels}), on the output latent graph without taking into consideration the potential pathways for compression. As such, the goal of this section is to outline effects of latent graph compression from two angles: (1) the impact of reducing the number of latent graph nodes via increases to the global node reduction factor $RF_G$, and (2) the impact of reducing the number of hidden channels $N_H$ stored on each latent graph node. It should be emphasized that both factors play into true compression achieved by the latent graph, as the total number of node degrees of freedom in $G_L$ is $(N_0^v/RF_G) \times N_H$, where $N_0^v$ is the number of nodes on the input graph $G_0$ (here, $N_0^v = 14476$). Since the primary focus and novelty of this work is tied to demonstrating the interpretability properties of the latent graph via the masked fields, the two compression angles described above are analyzed from the perspective of both reconstruction accuracy and identified structures in the masked fields. In other words, the focus here relates to how the GNN autoencoder provides the user the ability to access the ways in which latent graphs change to achieve flowfield compression.

\begin{figure}
    \centering
    \includegraphics[width=0.8\columnwidth]{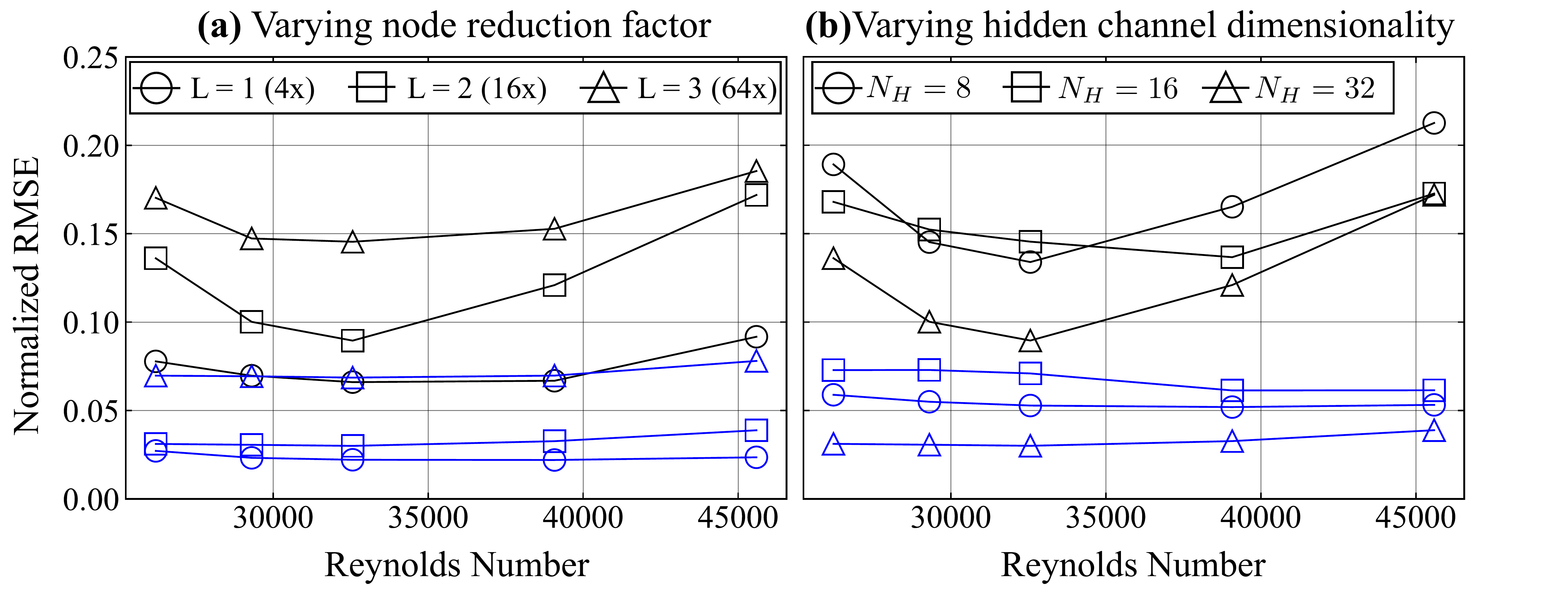}
    \caption{\textbf{(Left)} Normalized RMSE (see Eq.~\ref{eq:rmse}) as a function of Reynolds number for streamwise (black) and vertical (blue) velocity field reconstructions. Curves denote different node reduction factors obtained via increasing maximum graph level $L$ (refer to legend) for fixed $N_H=32$. \textbf{(Right)} Analogous to left, but different curves denote different $N_H$ values (refer to legend) for fixed $L=2$.}
    \label{fig:rmse}
\end{figure}

Figure.~\ref{fig:rmse} displays normalized root-mean squared errors (RMSE) at various Reynolds numbers corresponding to the dataset trajectories described in Sec.~\ref{sec:dataset}. The RMSE for a single trajectory (Reynolds number) is given by
\begin{equation}
    \label{eq:rmse}
    \text{RMSE} = \frac{\sqrt{ \frac{1}{M N_0^v} \sum_{i=1}^M \sum_{j=1}^{N_0^v} (x_{i,j} - \widetilde{x}_{i,j})^2 }}{u_{in}}, 
\end{equation}
where $x$ is a generic flowfield variable (e.g. $x = u_x$ for streamwise velocity and $x = u_y$ for vertical). In Eq.~\ref{eq:rmse}, $x_{i,j}$ and $\widetilde{x}_{i,j}$ denote the target and reconstructed flowfield variable respectively for the $i$-th snapshot and $j$-th node. Note that in Eq.~\ref{eq:rmse} the number of snapshots $M$ per trajectory varies, but the number of output graph nodes $N_0^v$ is fixed. To better interpret errors across the range of Reynolds numbers contained in the data, the baseline RMSE is normalized by the inlet freestream velocity $u_{in}$. 

Figure~\ref{fig:rmse}(a) shows the effect of increasing the node reduction factor on the normalized RMSE of Eq.~\ref{eq:rmse} for both streamwise and vertical velocity component reconstructions. To achieve increasingly higher amounts of node reduction in the latent graph, the maximum Top-K level in the GNN $L$ is adjusted with a fixed local reduction factor of $4$ between levels while freezing all other model hyperparameters (including the number of message passing layers). As a result, for the input graph containing $N_0^v=14476$ nodes, the latent graph in the $L=1$ model in Fig.~\ref{fig:rmse}(a) contains 4x fewer nodes, the $L=2$ model contains 16x fewer nodes, and the $L=3$ model 64x fewer nodes. Not that despite the fact that the number of nodes has been reduced by increasing $L$, the hidden channel dimensionality of all models in Fig.~\ref{fig:rmse}(a) is fixed to $N_H=32$ -- as such, only the $L=3$ model achieves true data compression in the sense of reduction in the total number of nodal degrees-of-freedom. 

Two immediate trends are apparent in Fig.~\ref{fig:rmse}(a) -- the first is that the errors are consistently higher for the streamwise component of velocity ($u_x$), and the second is that an increase in the node reduction factor via maximum Top-K level $L$ results in a vertical shift in the RMSE curves for both velocity components. The former trend is expected, as a majority of the flow contribution to the reattachment cycle dynamics is contained in the streamwise component. This is also evidenced by observing that the vertical component errors are largely insensitive to Reynolds number, wheras errors for streamwise velocity across all configurations shown in Fig.~\ref{fig:rmse} tend to increase with Reynolds number after about Re=32000. The latter trend conveys how it becomes increasingly more difficult for the decoder to recover the full flowfield as the number of nodes in the latent graph decreases. This is consistent with the fact that the same message passing scheme is used between all models (i.e. the MMP layer architecture is fixed as $L$ is increased), which in effect makes it more difficult for the decoder to populate flow information on the original nodes from the latent graph as the size of the latent graph decreases. Although not shown here, it is reasonable to expect that modifying the MMP layer design to compensate for the reduction in nodes, either in the form of adding additional coarsening levels or increasing the size of message passing blocks, would lessen the degree of the vertical curve shifts.

To complement the notion of the achieving compression by dropping the number of nodes, Fig.~\ref{fig:rmse}(b) displays error trends for models that instead modify the hidden channel dimensionality $N_H$ of the latent graph while fixing all other parameters. In particular, the autoencoders used to generate the curves shown in Fig.~\ref{fig:rmse}(b) come from taking the $L=2$ case in Fig.~\ref{fig:rmse}(a) (i.e. 16x reduction in the number of nodes) and adjusting the latent node dimensionality. As such, the $N_H=16$ case achieves a true compression factor of 2 and the $N_H=8$ case achieves one of 4. Although the model with the largest hidden channel dimensionality ($N_H=32$) tends to drop the RMSE curve over the tested Reynolds number range, the RMSE behavior at smaller values of $N_H$ is more complex -- for example, the error in vertical velocity component reconstructions is actually higher for $N_H=16$ when compared to $N_H=8$, although the same is generally not true for the more dominant streamwise velocity component. Despite this, the results in Fig.~\ref{fig:rmse} suggest that there is indeed an error-compression tradeoff from two angles: reduction in number of latent graph nodes and reduction in latent node dimensionality (hidden channels). As the scope of this work is directed towards demonstrating latent graph interpretability allowed by the architecture, further analysis of these error trends (i.e. in terms of compensating for error shifts with more message passing operations, increasing the number of input flow features, more rigorous hyperparameter testing, etc.) is omitted here and left for future reports.

\begin{figure}
    \centering
    \includegraphics[width=\columnwidth]{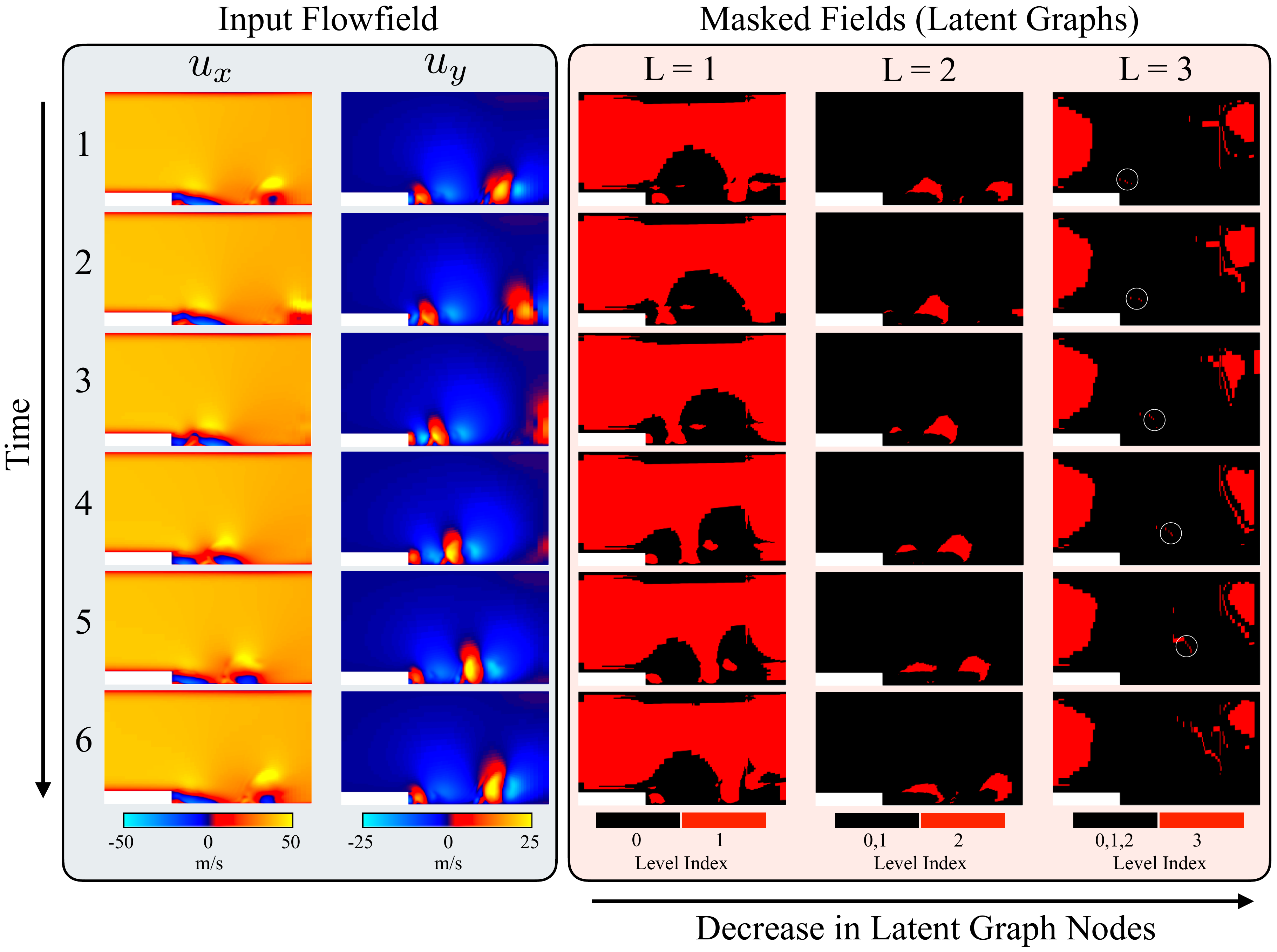}
    \caption{Snapshots from Trajectory 4 depicting reattachment cycle dynamics. First two columns display streamwise ($u_x$) and vertical ($u_y$) velocity fields, and next three columns display masked fields sourced from the same L=1, 2, and 3 models used in Fig.~\ref{fig:rmse}(a). Red regions in masked fields denote identified latent graph nodes. Time evolution proceeds from top-to-bottom, and successive snapshots shown here are separated by a time of $10^{-3}$ seconds. White circles in the $L=3$ case indicate additional small identified region in masked field.}
    \label{fig:time_evolution_mask}
\end{figure}

To better visualize the effect of node reduction on the latent graph, Fig.~\ref{fig:time_evolution_mask} shows the time evolution of masked fields (alongside input flowfields) during one reattachment cycle in trajectory 4 for the same $L=1$, $2$, and $3$ models plotted in Fig.~\ref{fig:rmse}(a). The figure illustrates how the adjacency matrix in the latent graph adapts to evolving flow structures during the shedding cycle, and also how increased reduction factor (compression) impacts the identified regions in the masked field to optimize reconstruction accuracy. The identified latent graph nodes (red regions in masked fields in Fig.~\ref{fig:time_evolution_mask}) are significantly impacted by the maximum graph level $L$ -- in other words, as the number of latent graph nodes decrease, the large-scale regions and flow features in physical space identified by the masked fields change. For example, the $L=1$ case (node reduction factor of 4) identifies the inlet and outlet boundaries, a majority of the freestream region, and also the structure of propagating recirculation zones in the flow. On the other hand, as a result of increasing the node reduction factor to 16, the latent graph in the $L=2$ case eliminates concentration in the freestream regions, and instead focuses on the shedding structures in the step cavity (the physical significance of these identified regions were oultined in Sec.~\ref{sec:topk_levels}, for which the same $L=2$ model was used). It is clear from Fig.~\ref{fig:time_evolution_mask} that identified structures propagate in accordance with the shedding cycle frequency. Note that although the number of latent graph nodes has dropped by a factor of 4x when moving from the $L=1$ to the $L=2$ case, there is a disproportionate decrease in the "coverage" of the masked fields in physical space due to the fact that the mesh resolution in the near-step region is much higher. Interestingly, the $L=3$ case -- the model that achieves a node reduction of 64x and achieves a true compression in the degrees of freedom by a factor of 4 -- isolates only the inlet and outlet boundary regions. Although a very small time-evolving structure near the step cavity is identified to capture the unsteadiness in the reattachment cycle (indicated by white circles in Fig.~\ref{fig:time_evolution_mask}), the latent graph connectivity in the $L=2$ case is largely insensitive to the evolving flow patterns. This implies that in cases of greater latent graph reduction, the Top-K mechanism isolates the inlet and exit boundary conditions as opposed to the step cavity region as the primary mechanism for flow reconstruction. 

\begin{figure}
    \centering
    \includegraphics[width=0.7\columnwidth]{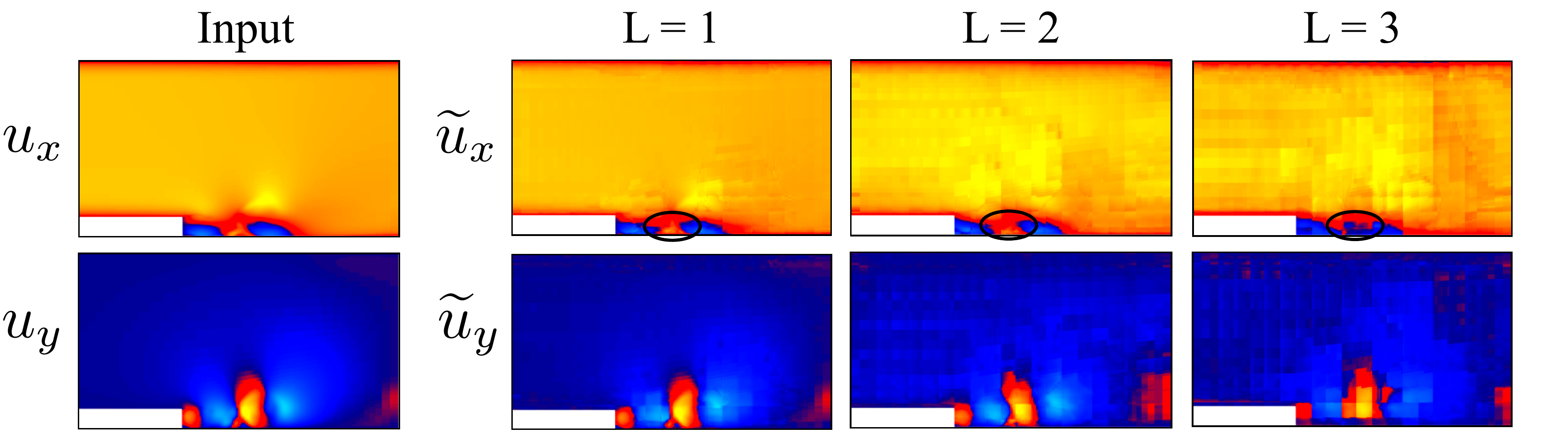}
    \caption{Flowfield reconstructions for the same L=1, 2, and 3 models utilized in Figs~\ref{fig:rmse}(a) and Fig.~\ref{fig:time_evolution_mask}. Input flowfield corresponds to snapshot index 4 in Fig.~\ref{fig:time_evolution_mask}. Colorbars are also consistent with Fig.~\ref{fig:time_evolution_mask}.}
    \label{fig:reconstruction_levels}
\end{figure}

Figure~\ref{fig:reconstruction_levels} shows how these masked fields translate to flowfield reconstructions of varying accuracy in accordance with the amount of node reduction provided by latent graph. More specifically, shown in Fig.~\ref{fig:reconstruction_levels} are reconstructions for streamwise and vertical velocity components corresponding to snapshot 4 in Fig.~\ref{fig:time_evolution_mask}, which depicts a moment just after the pinch-off phenomenon has occurred. Upon inspection, it is clear that the reconstruction accuracy -- especially in terms of the presence of non-physical flow artifacts (e.g. discontinuities due to interpolation) -- begins to drop as the maximum level $L$ increases. Despite this, although deviation in the freestream region appears in the $L=2$ model, the large-scale structures near the step are captured well. However, the $L=3$ model begins to see significant deterioration in the reconstructed flow features near step -- in particular, the gap between pockets of negative streamwise velocity components is not captured, as indicated by the black circled regions in the respective figures. Despite this, it is motivating that the latent graph even in the $L=3$ case is able to recover the general structure of the BFS flow features in light of the fact that the identified nodes are largely concentrated away from the step and towards the boundaries. 

Overall, the trends observed in Fig.~\ref{fig:reconstruction_levels} reflect those observed in Fig.~\ref{fig:rmse}(a), in that there is a direct deterioration in reconstruction accuracy as the amount of node reduction is increased. Although the reconstruction accuracy (particularly for the $L=3$ case in Fig.~\ref{fig:reconstruction_levels}) leaves something to be desired, it is emphasized that the advantage and objective of the autoencoding framework used here is the ability for the user to access and interpret the discovered latent graph through the visualization of time-evolving masked fields, as shown in Fig.~\ref{fig:time_evolution_mask}. As mentioned above, a natural step forward is to explore the impact of including additional message passing operations in the MMP layer, as well as graph-based filtering operations, to offset the loss of reconstruction accuracy due to increased levels of node reduction. Such aspects will be reported elsewhere. 

\section{Conclusion}
\label{sec:conclusion}
A graph autoencoder leveraging a combination of adaptive Top-K pooling layers and multiscale message passing (MMP) layers was introduced in this work. The pooling operation is interpreted as an adaptive sampling mechanism -- stacking such layers results in a latent space (here referred to as a latent graph) that can readily be interpreted by visualizing the identified nodes through the construction of a masked field. The goal of the MMP layers is to then redistribute information contained in the masked field to the rest of the domain, resulting in a reconstruction of the original input graph. Alongside providing details and nuances of the new architecture, the primary goal here was to showcase the interpretability properties of generated latent spaces (or latent graphs) produced in the encoding stage in fluid dynamics applications. To this end, datasets for training and evaluation purposes were generated using large-eddy simulations in a backward-facing step (BFS) flow configuration with an \verb|OpenFOAM|-based flow solver at high Reynolds numbers. Using this dataset as testbed, important aspects related to GNN architecture design, physical interpretation of latent spaces, and flowfield reconstruction quality were analyzed. 

From the angle of architecture design and to motivate the need for the MMP layer, the effect of coarsening operations present in the encoder and decoder message passing operations were analyzed. This analysis was conducted by comparing outputs of three model variations: (1) a GNN with no coarsening operations in the MMP layer (i.e. no multiscale message passing), (2) a GNN with coarsening operations present only in the decoder, and (3) a GNN with coarsening operations present in both encoder and decoder. In the end, the first model type resulted in a failure of the GNN to recover full flowfield information from the masked fields -- incorporating coarsening operations via models 2 and 3 resulted in much more accurate and complete reconstructions, illuminating the advantages provided by multiscale message passing operations. The variation in architecture type directly impacted the physical nature of the masked fields: interestingly, the inclusion of coarsening operations in the encoding stage (model 3) was found to effectively trades physical space coherency and interpretability in the identified latent graph for improved reconstruction accuracy.

Physical interpretation of the masked fields at different stages of the reattachment cycle revealed how the latent graphs directly identify time-evolving coherent structures -- in most cases, even in light of changes to the Top-K hierarchy (i.e. modification to the maximum level $L$ for a fixed reduction factor), the nodes sampled in the reduction procedure were strongly correlated with recirculation zones of varying strengths in the flow. Flowfield compression potential was then assessed from two angles: by increasing node reduction factors through variations in maximum Top-K level, and by decreasing the hidden channel dimensionality of latent  graph nodes. From both perspectives, it was found that the average reconstruction errors understandably increased across the board for greater compression factors. Although reconstruction qualities were imperfect in the higher-compression models, the advantage of this autoencoding approach is that it provides the user insight into how the greater levels of compression are achieved via visualization of the time-evolving masked fields. More specifically, it was found that the model achieving the highest amount of node reduction placed more focus on the inflow and outflow boundaries in the domain than other models. Additionally, when considering errors for individual flowfield variables at a given level of compression, reconstruction errors in the form of an RMSE measure were found to be much more insensitive to Reynolds number in the vertical velocity component when compared to streamwise velocity counterparts. 

Because the scope of this work was tied primarily to demonstrating the latent space interpretability provided by the GNN autoencoder, there are many avenues for future work. Concerning the architecture parameters, investigating the effects of increased message passing and coarsening operations within the MMP layer is warranted, as it may potentially alleviate issues related to the increase in reconstruction error found in higher-compression models. Additionally, geometry extrapolation capability was unexplored in this work -- assessment of latent graph properties in different geometric configurations is a promising direction to solidify the general applicability of the method. From the reduced-order modeling perspective, the concept of masked field evolution reveals new strategies for interpretable data-based surrogate modeling of complex fluid flow phenomena (e.g. by establishing quantitative connections between flowfield dynamics and masked field evolution). All of these aspects are actively being pursued, and will be reported in future studies.

\section{Acknowledgements} 
This research used resources of the Argonne Leadership Computing Facility, which is a U.S. Department of Energy Office of Science User Facility operated under contract DE-AC02-06CH11357. RM acknowledges funding support from ASCR for DOE-FOA-2493 ``Data-intensive scientific machine learning".

% \bibliographystyle{unsrt}
% \printbibliography
\bibliography{refs}

\end{document}